\documentclass{egpubl}
\usepackage[normalem]{ulem} %

\usepackage{egsgp19}

\SpecialIssuePaper         %
\usepackage[T1]{fontenc}
\usepackage{dfadobe}  

\usepackage{cite}  %
\BibtexOrBiblatex
\electronicVersion
\PrintedOrElectronic
\ifpdf \usepackage[pdftex]{graphicx} \pdfcompresslevel=9
\else \usepackage[dvips]{graphicx} \fi

\usepackage{egweblnk} 

\usepackage{times}
\usepackage{epsfig}
\usepackage{graphicx}
\usepackage{amsmath}
\usepackage{amssymb}
\usepackage{booktabs}
\usepackage{bm}
\usepackage{ulem}
\usepackage{subcaption}
\usepackage{xcolor}
\usepackage{gensymb}
\usepackage{hyperref}

\DeclareOldFontCommand{\bf}{\normalfont\bfseries}{\textbf}
\newcommand{\myparagraph}[1]{\vspace{3pt}\noindent{\bf #1}}

\definecolor{aqua}{rgb}{0.0, 0.48, 0.65}

\newcommand{\final}[1]{\textcolor{black}{ #1}}

\begin{document}

\title{Unsupervised cycle-consistent deformation for shape matching}

\author[T. Groueix \& M. Fisher \& V. G. Kim \& B. C. Russell \& M. Aubry]{
Thibault Groueix$^{1}$, Matthew Fisher$^2$, Vladimir G. Kim$^2$, Bryan C. Russell$^2$, Mathieu Aubry$^1$\\
$^1$LIGM (UMR 8049), \'Ecole des Ponts, UPE, $^2$Adobe Research\\
{ \url{http://imagine.enpc.fr/~groueixt/cycleconsistentdeformations/}}\\
}

\maketitle
\begin{abstract}
We propose a self-supervised approach to deep surface deformation. Given a pair of shapes, our algorithm directly predicts a parametric transformation from one shape to the other respecting correspondences. Our insight is to use cycle-consistency to define a notion of good correspondences in groups of objects and use it as a supervisory signal to train our network. Our method does not rely on a template, assume near isometric deformations or rely on point-correspondence supervision. We demonstrate the efficacy of our approach by using it to transfer segmentation across shapes. We show, on Shapenet, that our approach is competitive with comparable state-of-the-art methods when annotated training data is readily available, but outperforms them by a large margin in the few-shot segmentation scenario.
\end{abstract}

\section{Introduction}

\begin{figure}[t]
\centering
 \includegraphics[width=0.8\linewidth]{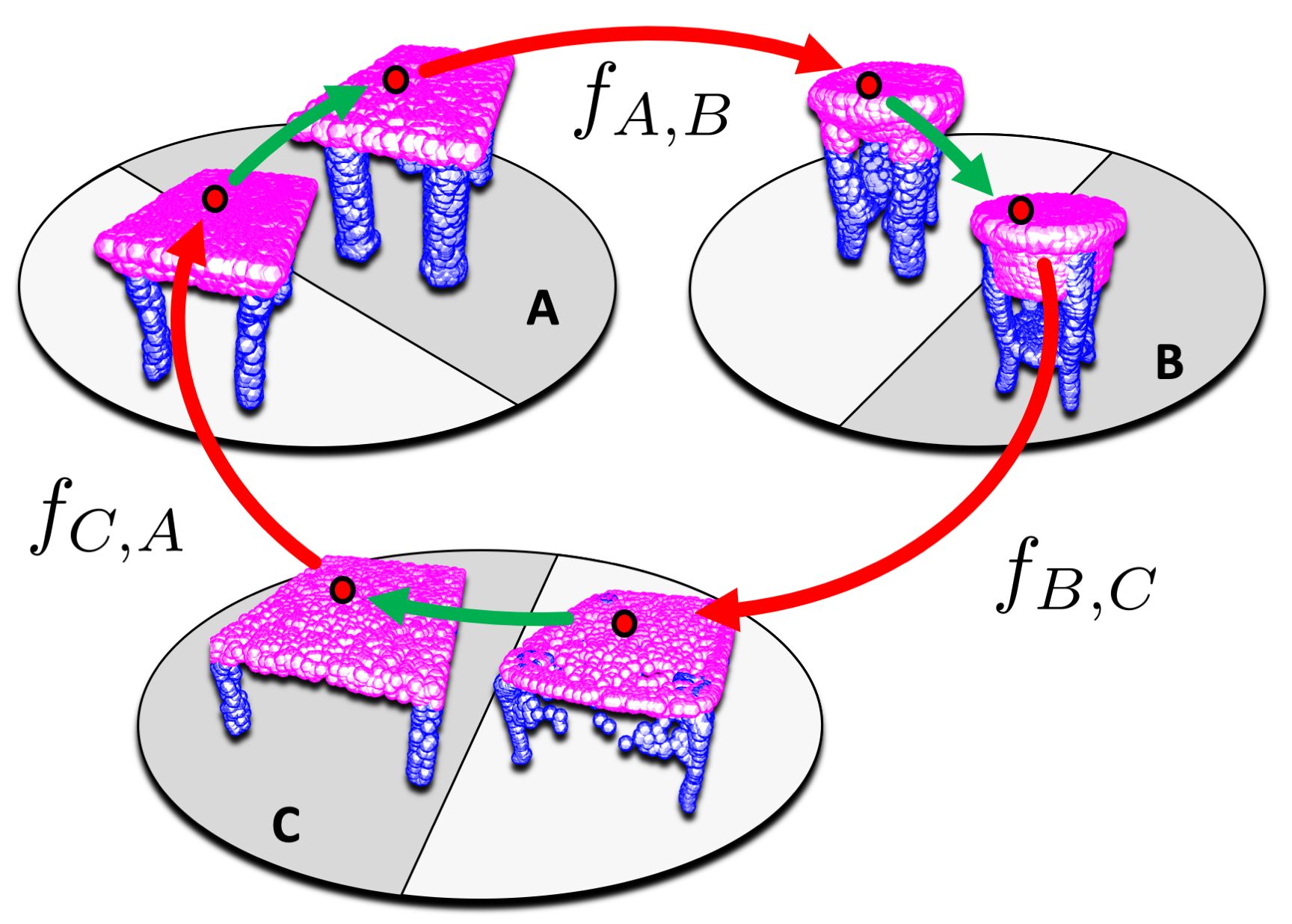}
\caption{\textbf{Shape deformation with cycle-consistency.} Our approach takes a pair $(A, B)$ of pointclouds as input and predicts a deformation of A into B. During training, a cycle-consistent loss on a shape triplet (A, B, C) allows the method to learn semantically consistent deformations $f_{A,B}$, $f_{B,C}$, $f_{C,A}$ without any priors. Red arrows represent the learned shape deformation function and green arrows indicate the projection of the deformed shape onto the nearest point on the surface of the target shape.
}
  \label{fig:teaser}
\end{figure}

Large collections of 3D models enable data-driven techniques for interactive geometry modeling, shape synthesis, image-based reconstruction, and shape completion~\cite{Mitra14}. Many of these techniques require the collection to have additional surface annotations such as segmentation into functional~\cite{Yi16} or geometric~\cite{Li2018} parts.
The notion of parts and their granularity can vary significantly across different tasks, so many novel applications require new types of annotations~\cite{mo2018partnet,yu2019partnet,wang2019shape2motion}. 
Deep learning algorithms have recently achieved state-of-the-art in automatically predicting such
surface annotations~\cite{qi2016pointnet,qi2017pointnetplusplus,Wang18}. However, they typically require a significant number of training examples for every shape category, which limits their applicability, and bears significant start-up cost in introducing a new type of annotation. In this work, we propose a new deep learning approach which leverages large non-annotated object collections to perform few-shot segmentation.

We rely on the idea to use shape matching to transfer labels from similar examples. This approach has been shown to be robust in extreme ``few-shot'' learning scenarios~\cite{Yi16} and can work robustly even in heterogeneous datasets as long as labeled models roughly span all the shape variations. The few-shots segmentation problem then amount to the fundamental problem of identifying correspondences between shapes. 
There is a vast amount of work on shape matching, which can be roughly separated in two trends: (i) classical optimization based approaches; (ii) recent approaches where correspondences are directly predicted by a neural network. 

Traditional, optimization-based methods such as iterative closest point (ICP) algorithm, are fast and effective with good initial guesses and few degrees of freedom (e.g., a rigid motion)~\cite{Rusinkiewicz01}. More flexible correspondence algorithms for dissimilar models usually require significantly more compute time to optimize for larger number of degrees of freedom~\cite{Brown07,Kim11,Chen15}. Since directly matching dissimilar shapes poses significant challenges, these methods often rely on joint analysis of the entire collection~\cite{Kim12}, leveraging cycle consistency priors during optimization~\cite{Huang2013,nguyen2011optimization}. These joint correspondence estimation methods tend to be very compute heavy and as new models are added to the collection, the entire optimization needs to be repeated. We thus turned to deep learning-based approaches.

Indeed, with the recent advances in neural networks for geometry analysis, learning-based methods have been proposed to address the matching problem. Of particular interest to us is the method of Groueix et al.~\cite{groueix2018b}, which demonstrate that one can learn how to deform a human body template to the target point cloud, even without correspondence supervision. In their approach, the target point cloud is encoded into a latent descriptor space (via PointNet encoder~\cite{qi2016pointnet}), and then the deformation network takes the target descriptor and a point on the template, and maps the point to new position so that it aligns to the target. 
This approach is efficient, since it only requires a forward pass through a network. It also has the benefit of holistic understanding of shape deformations, since the same neural network is trained for all models in the input collection. However, it has to be trained specifically for each template, limiting this method to analysis of geometrically and topologically similar shape collections, such as human bodies.
If such a template is not available, one can pick a very generic shape (e.g., a sphere) and still obtain some correspondences via the intermediate domain~\cite{groueix2018}. However, as we will show, the quality of the correspondences will degrade significantly as shapes deviate from that domain. 

In this work we propose a novel neural network architecture that learns to match shapes directly, without relying on a pre-defined template, by learning to predict deformations that aligns points on the source shape to points on the target. Note that the transformation can be much more complex than a rigid transformation, and that the space of meaningful transformation is defined implicitly by the (unlabelled) training data. We encode both source and target shapes and then predict the deformed position for every point on the source conditioned on these two codes, unlike prior work that use a fixed template common to all the shapes. 
We show that the results obtained can be greatly improved if the network is trained not only with a reconstruction loss, which encourages it to deform the source shape into the target shape, but also using a cycle consistency loss. Indeed a deformation which respects correspondences should be consistent between pairs of shapes \textit{i.e.}, the deformation from A to B should be the inverse of the deformation from B to A
\final{. More generally, in larger cycles of shapes $[A_1, ..., A_i, .., A_N]$, global consistency is achieved if the composition of the N successive mappings from $A_i$ to $A_{i+1}$ is identity.} This new consistency loss used during training can be seen as playing a role similar to the global consistency objective used in optimization-based approaches.
Finally, our network is trained in an self-supervised manner using only shape reconstruction and cycle consistency losses. 

We demonstrate the effectiveness of our approach for shape matching by propagating segmentations in a few-shot learning setting on the ShapeNet part dataset~\cite{Yi16}. 
We first show that in this extreme case with very few training examples, PointNet~\cite{qi2016pointnet}, a strongly supervised method, fails to generalize. 
Then, we propose several strategies for picking source shapes and propagate the signal from them, using our predicted correspondences. 
We demonstrate that even with a simple strategy, such as picking the source with smallest Chamfer distance, our method is better at transferring segmentations than other fast correspondence techniques such as ICP with rigid transformation and a prior learning-based method that aligns sphere and plane templates~\cite{groueix2018}. 

\begin{figure*}[!ht]
\centering
\begin{subfigure}[b]{0.48\linewidth}
\centering
 \includegraphics[width=\linewidth]{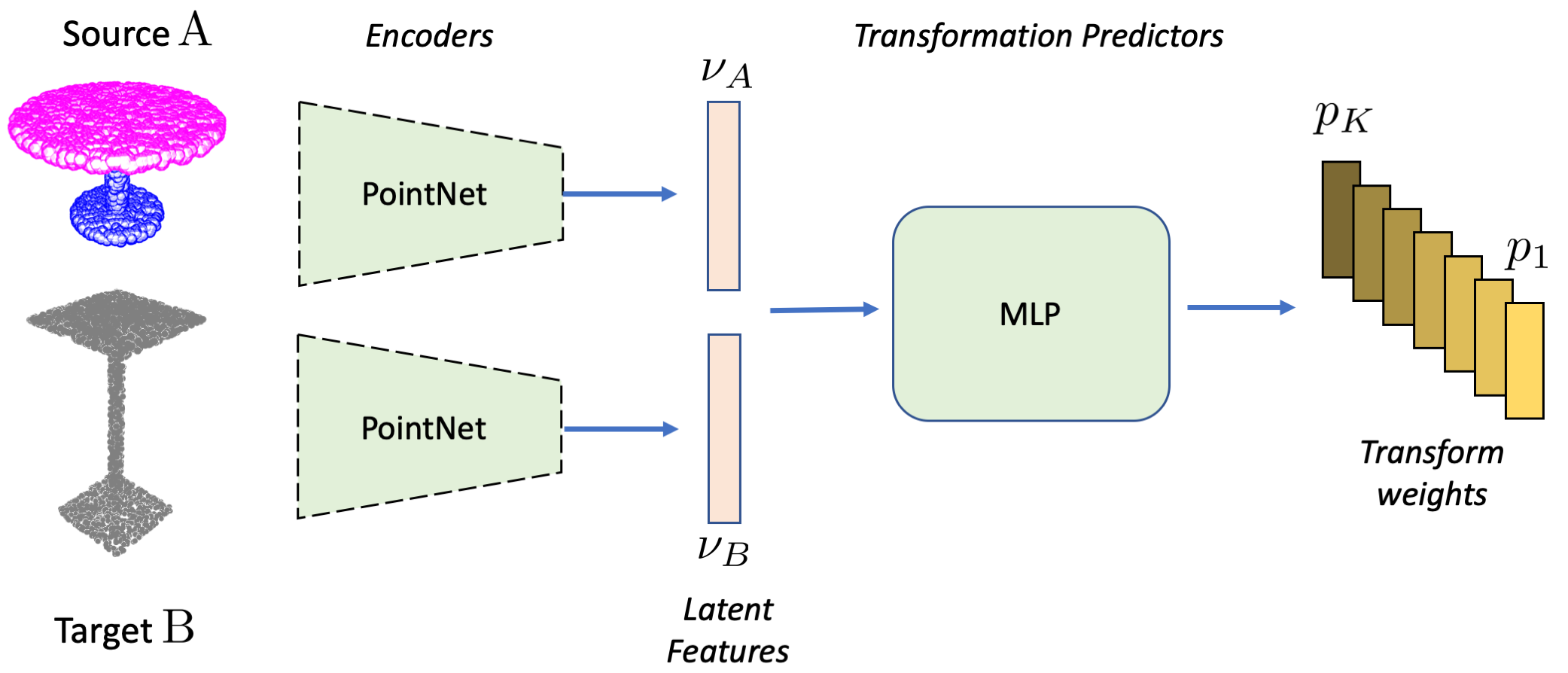}
\caption{Parameter prediction network. \label{fig:meta_network}}
\end{subfigure}
~\vline~
\begin{subfigure}[b]{0.48\linewidth}
\centering
 \includegraphics[width=\linewidth]{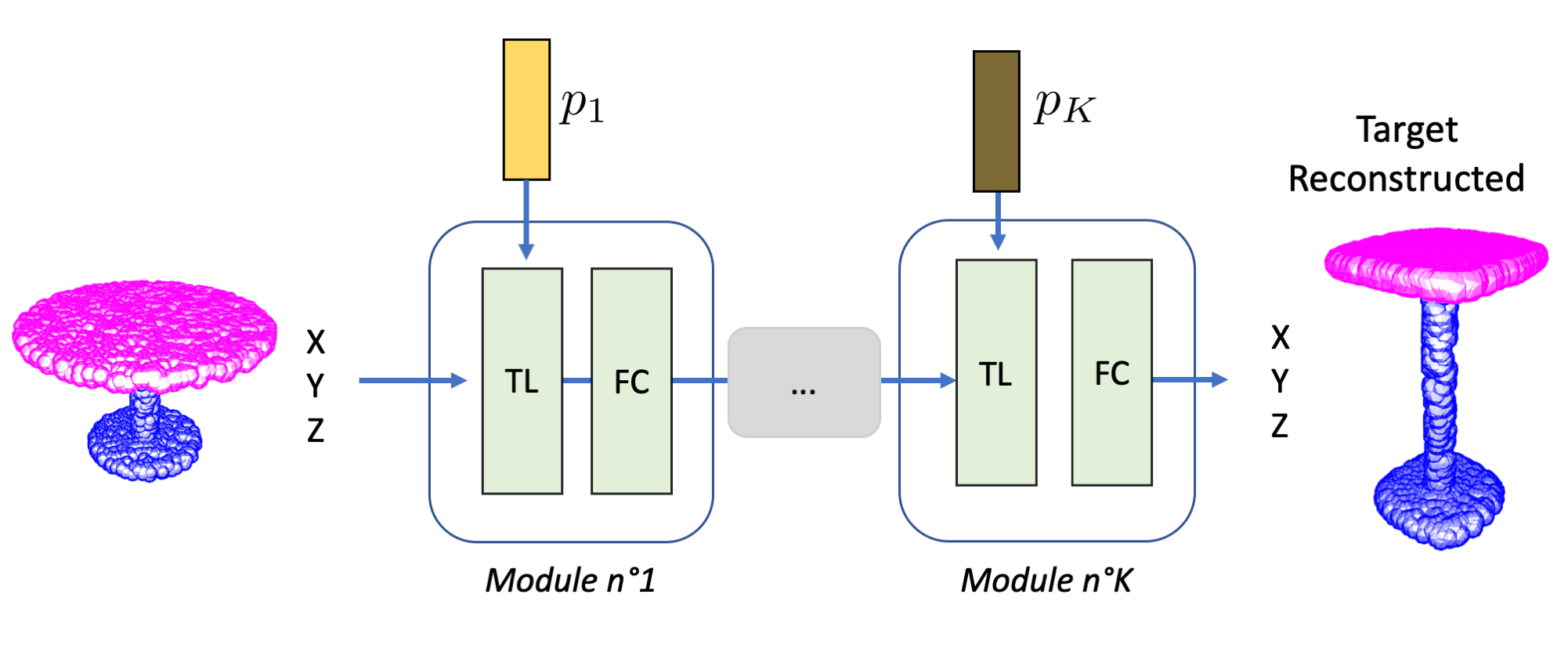}
\caption{Deformation network. \label{fig:deform_network}}
\end{subfigure}
\caption{\textbf{Shape Deformation approach.} Our methods take as input a pair (source $A$, target $B$) of shapes and aims at predicting the deformation of $A$ in $B$.  In \textbf{(a)}, $A$ and $B$ are encoded with Pointnets~\cite{qi2016pointnet} into a latent feature vector, from which an MLP predicts transformation parameters, used in \textbf{(b)} to deform $A$ into $B$, by stacking Transformation Layers \textbf{(TL)} and Fully-Connected Layers \textbf{(FC)}.
}
  \label{fig:overview}
  \vspace*{-4mm}
\end{figure*}

\section{Related Work}
Shape matching is a long-standing problem in shape analysis~\cite{vankaick11correspsurvey}. It is often done explicitly, by deforming a source shape to a target~\cite{Rusinkiewicz01,Brown07,li08global,HAWG2008NRUID,zhang_sgp08}, or implicitly, by mapping points~\cite{Kim11,Chen15,OvsjanikovMMG10,Bronstein1168} or functions~\cite{Ovsjanikov2012,Ren2018,Ezuz2017} on one shape to another. The deformation-based methods typically aim to minimize the amount of distortion introduced by the deformation, and the mapping-based approaches often assume that shapes to be near-isometric. Both assumptions do not hold for very dissimilar shapes. 

To address this challenge, some prior methods leverage additional context of the entire shape collection in a joint optimization~\cite{Kim12,nguyen2011optimization}. These techniques often use cycle-consistency as additional cue~\cite{Huang2012,Huang2013,Rustamov2013}. This, enables estimating correspondences even between dissimilar objects by mapping via intermediate shapes. 
While these traditional optimization techniques are very powerful, non-rigid matching involves optimizing for many degrees of freedom with complex non-convex objective functions, and takes minutes or hours. To make matters worse, joint analysis usually scales in a super-linear manner with number of models, and if a new shape is added to a collection, the entire optimization needs to be repeated. 

Recently, learning-based correspondence techniques were used to address these limitations. They are fast, typically only requiring a forward pass through a neural network, and they enable joint analysis of a collection of shapes, since multiple shapes are typically used during training. Descriptor-based methods embed each shape point into some high-dimensional space, where corresponding points are embedded nearby~\cite{Huang18,BoscainiMRB16,wei2016dense}. In most cases, however, a more holistic mapping for the entire shape is often preferred, since it is more capable of preserving the intrinsic shape structure. Litany et al.~\cite{LitanyRRBB17} use a deep neural network to predict a soft inter-surface mapping a common representation used in functional map framework. Groueix et al.~\cite{groueix2018b} propose to train a network that predicts a deformation for each point on a template. A similar method that uses planes or spheres can be used in case such a template is not available~\cite{groueix2018}. These techniques struggle with diverse shape collections when matched shapes have very different topology and geometry. Instead, we propose a method that takes both source and target shape as input and infers the mapping. We also propose a novel regularization term favoring cycle-consistency when mapping across multiple shapes in the collection. \final{A similar cycle-consistency loss for training deep networks to predict correspondences between images of different instances of objects from the same category has recently been used in~\cite{zhou2016learning}. In this work, views rendered from different viewpoints from a 3D model were used to avoid the trivial identity flow solution, but no correspondence between 3D shapes was predicted.}

We demonstrate the value of our method for few-shot segmentation transfer. While many techniques have been developed for strongly supervised mesh segmentation~\cite{qi2016pointnet,qi2017pointnetplusplus,Wang18,Li2018,Kalogerakis:2017:ShapePFCN,Kalogerakis:2010:labelMeshes}, they typically rely on many training examples and fail in a few-shot scenarios (see Table~\ref{tab:fewshot}). In these cases, some framework propose to rely on propagating annotations from most similar annotated shapes via global or local shape matching~\cite{Yi16}. In fact, it is common for correspondence techniques to be evaluated and used for transferring various signals between shapes~\cite{Ovsjanikov2012,Kim11,Azencot2017,ChangFGHHLSSSSX15}.

\section{Learning asymmetric cycle-consistent shape matching}

\label{sec:main}

We address the surface matching problem by training a model that takes as inputs a source shape, a target shape, and a point on the source shape and generates the corresponding point on the target shape. As pointed out in Groueix et al.~\cite{groueix2018b}, a learnable model allows for efficient surface matching, which is in contrast to approaches requiring optimization over a collection of pairwise shape matches~\cite{nguyen2011optimization}. 

We assume that shapes are represented as point sets sampled from the shapes' surface. Given point sets $A$ and $B$, our goal is to learn a mapping function $f_{A,B}$ that takes a 3D point $\mathbf{p}\in A$ to its corresponding point $\mathbf{q}\in B$. If $f$ is a function on points and A a set of points, we denote by $f(A)$ the set $\{f(\mathbf{p}), \forall \mathbf{p}\in A\}$. %

First, building on work on unsupervised template-based shape correspondence~\cite{groueix2018b} we use a Chamfer loss to minimize the distance between deformed source $f_{A,B}(A)$ and the target $B$. Unlike prior work, however, we do not assume that all of our shapes are derived from the same template and directly predict template-free correspondences between pairs of shapes.%

Second, we seek to leverage the success of cycle consistency, which has been used in shape collection optimization~\cite{nguyen2011optimization} and more recently in self-supervised learning~\cite{CycleGAN2017}, during training of our learnable mapping function. 
Formally, for $N$ shapes $X_1,\dots,X_N$ that are assumed to be put into correspondence, we enforce that the learnable mapping function $f_{A,B}$ satisfies,
\begin{equation}
  \forall \mathbf{p}\in X_1, \  f_{X_1,X_2}\circ \dots \circ  f_{X_{N-1},X_N}\circ  f_{X_{N},X_1}(\mathbf{p}) = \mathbf{p}.
\end{equation}
We use cycle-consistency training losses for cycles of lengths two and three as it implies consistency for cycles of any length~\cite{nguyen2011optimization}. We visualize our cycle-consistency loss in Figure~\ref{fig:teaser}.

\section{Approach}

We describe our learnable mapping function $f_{A,B}$, implemented as a two-stage neural network, in Section~\ref{sec:architecture}, our training losses in Section~\ref{sec:losses}, and application to segmentation in Section~\ref{sec:segmentation}.

\subsection{Architecture}
\label{sec:architecture}

The architecture of our shape transformation model from a source shape A to a target shape B is visualized in Figure~\ref{fig:overview} and can be separated into two parts: (a) a parameter prediction network which outputs transformation parameters given the two shapes~(Figure~\ref{fig:overview}a); (b) a deformation network that transforms the first shape into the second one using the predicted parameters~(Figure~\ref{fig:overview}b). We now describe these two components.

To predict transformation parameters, A and B are first passed into two independent PointNet networks~\cite{qi2016pointnet} leading to feature encodings $v_A$ and  $v_B$ \final{of size 512}. The resulting concatenated descriptor $v_{AB} = [v_A, v_B]$ contains information about the pair (A, B). 
A multilayer perceptron (MLP) then predicts transformation parameters vectors $p_1, \cdots ,p_K$ from this concatenated feature.%

 The deformation network (Figure \ref{fig:overview}b) takes a surface point in $\mathbb{R}^3$ and outputs the associated deformed point. The network is composed of $K$ modules each with the same architecture. Let's call $x_{k-1}$ the input of module $k$ and $x_k$ its output. The operation computed by this module is:
\begin{equation}
    x_k=A_k\left(W_k\left(s_k\cdot x_{k-1}+b_k\right)\right),
\label{eq:eg2} 
\end{equation}
where $W_k$ is the matrix of parameters of a fully-connected layer \final{in $\mathbb{R}^{64x64}$}, "$\cdot$" refers to the Hadamard (term to term) product, $A_k$ is the activation function for module $k$ and $[s_k, b_k]=p_k$ are the transformation parameters, \final{both in in $\mathbb{R}^{64}$}, corresponding to a scale and a bias in each dimension. Note that this is similar to the architecture of the T-net modules in \cite{qi2016pointnet, jaderberg2015spatial}, but using fewer predicted parameters. \final{Also note that equation~\ref{eq:eg2} is differentiable, which enables the two sub-networks to be trained jointly in an end-to-end fashion.}
In all of our experiments we used $K=7$ modules, 64 dimensions for each intermediary feature and ReLU activations for all but the last layer, for which we used a hyperbolic tangent. \final{We train for 500 epochs with Adam~\cite{kingma2014adam} starting with a learning rate of $0.01$ divided by 10 after 400 epochs.}

\subsection{Training Losses}
\label{sec:losses}

We train our deformation by minimizing the \final{\sout{weighted}} sum over several components: a loss enforcing cycle consistency $L_{\mathbf{Cy}}$, Chamfer distance loss $L_\mathbf{Ch}$, and a self reconstruction loss $L_\mathbf{SR}$ :
$$L_\mathbf{total} = L_\mathbf{SR}+ L_\mathbf{Ch} + L_{\mathbf{Cy}}  $$
We only use the self-reconstruction loss to stabilize the beginning of the training and disable it after 30 epochs to focus on cycle consistency and reconstruction losses. 
We train all parameters in our network by sampling triplets $(A, B, C)$ of shapes which are needed by our 3-cycle consistency and enforcing all other losses on all the associated deformations. We first explain how we sampled these triplets, then detail the different terms of our loss.

\subsubsection{Training shape sampling}
\label{sec:sampling}

For our cycle-consistency loss, we require a valid mapping across shape triplet (A, B, C). As different shape categories may have different topologies, we train category-specific networks. Furthermore, as there may be topological changes within a single category, for shape A, we randomly sample shapes B and C from the K nearest neighbors of A under chamfer distance. We take $K=20$ and demonstrate in the ablation study the superiority of this approach over random sampling of shape triplets.

We apply data augmentation $\psi$ on each sampled shape in this order : a random rotation around the $Z$ axis of a random angle between $-40\degree$ and $40\degree$, an anisotropic scaling of random scale between $0.75$ and $1.25$, a bounding box normalization, and a small random translation below 0.03.

\subsubsection{Cycle-consistency loss}
The cycle consistency loss is based on the intuition that a point deformed through any cycle of deformations should be mapped back to itself. One way to enforce consistency would be to compute composite functions, for two shapes $X$ and $Y$ minimizing $\| \mathbf{p}-f_{Y,X}\circ f_{X,Y}(\mathbf{p})\|$ for all $\mathbf{p}$ in $X$. However $ f_{X,Y}(\mathbf{p})$ is typically not an element of $Y$, and computing $f_{Y,X}\circ f_{X,Y}(\mathbf{p})$ would thus require computing the deformations $f_{Y,X}$ of other points than the points of $Y$. To avoid this, we consider instead projections of the deformed shapes to the target shapes. More precisely, we define the shape projection operator $\pi$
\begin{equation}
    \pi_X(\mathbf{p})=\text{argmin}_{\mathbf{q}\in X}\|\mathbf{p}-\mathbf{q}\|
\end{equation}
and enforce 2-cycle consistency between $X$ and $Y$ by minimizing
 \begin{equation}
    Cy_2(X, Y) =  
    \frac{1}{\left|X\right|}\sum_{\mathbf{p}\in X} \left|\mathbf{p} - f_{Y,X}\circ \pi_Y\circ f_{X,Y}(\mathbf{p}) \right|_{2}
\end{equation}
and cycle consistency for the $(X,Y,Z)$ cycle by minimizing 
\begin{equation}
    Cy_3(X, Y, Z) = 
    \frac{1}{\left|X\right|}\sum_{\mathbf{p}\in X} \left|\mathbf{p} - f_{Z,X}\circ  \pi_Z \circ f_{Y,Z} \circ \pi_Y \circ f_{X,Y}(\mathbf{p})\right|_{2}
\end{equation}
Our full cycle-consistency loss $L_{\mathbf{Cy}}$ is simply defined by summing over possible all possible two and three cycles using a sampled triplet (A, B, C).
\begin{equation}
L_{\mathbf{Cy}} = \sum_{X,Y,Z \in \{A,B,C\} s.t. \{X,Y,Z\}=\{A,B,C\}} Cy_2(X, Y) + Cy_3(X,Y,Z)
\end{equation}
\final{Enforcing 2- and 3-cycle consistency implies consistency for any cycle~\cite{nguyen2011optimization}.}

\subsubsection{Reconstruction loss}

\final{As discussed in section \ref{sec:main}, we want to enforce that every point in the target shape is well reconstructed, but not necessarily that any point in the source shape is mapped to the target shape, in case some part appear in the source and not the target. We thus used asymmetric Chamfer distance to quantify how well the network has generated the target shape.}  More precisely, given a pair of shapes (X,Y), the asymmetric chamfer $Ch(X, Y)$ computes the average distance between a point $\mathbf{q}\in Y$ and its nearest neighbor in $X$. 
\begin{equation}
    Ch(X, Y) =  %
    \frac{1}{\left|X\right|}\sum_{\mathbf{q}\in Y} \min_{\mathbf{p}\in X} \left\| p - q \right\|_{2}.
    \label{eq:chamfer}
\end{equation}

Given a training triplet $(A,B,C)$, we define the reconstruction loss by summing the asymmetric chamfer loss on all 6 possible (source, target) couples. %
\begin{equation}
L_\mathbf{Ch} = \sum_{X,Y \in \{(A,B), (A, C), (B, C)\}}Ch(f_{X,Y}(X), Y) + Ch(f_{Y,X}(Y), X)
\end{equation}

If segmentation is available for the training shapes, we can compute the distance in equation \ref{eq:chamfer} on each segment independently, which would add supervision on the correspondences. We of course do not use such labels for our few-shot learning experiments, but show in Table \ref{tab:supervisedsegmentation} it can be used if available to slightly boost our results.

\subsubsection{Self-reconstruction loss}
We can fully supervise the deformation by manually deforming a shape with a known transformation. We found such a supervision was helpful to stabilize and speed up the beginning of our training. Concretely, we sampled deformations $\psi$ similar to what we did for data augmentation (described above in \ref{sec:sampling}) by composing (1) a rotation, (2) an anisotropic scaling, and (3) a rescaling to a centered bounding box. %
Given a transformation $\psi$, we compute the average distance between the two images of a point $\mathbf{p}\in A$ under $\psi$ and the predicted mapping function $f_{A,\psi(A)}$.
\begin{equation}
    SR(A, \psi) =  \frac{1}{\left|A\right|}\sum_{\mathbf{p}\in A}  \left\| f_{A,\psi(A)}(p) - \psi(p)  \right\|_{2}
\end{equation}
Our corresponding self-reconstruction loss $L_\mathbf{SR}$ is the sum of this loss for each of the three point clouds in the triplet (A, B, C) with different random transformations.
\begin{equation}
    L_\mathbf{SR}  = SR(A, \psi) + SR(B, \psi') + SR(C, \psi'')
\end{equation}

\subsection{Application to segmentation}
\label{sec:segmentation}

Learning a deformation between two shapes provides an intuitive method to transfer label information, such as a part segmentation, from a labeled shape to an unlabeled one. In this formulation, we assume we are given a (small) number of labeled shapes, and seek to label each point on an unlabeled test shape. This requires us to decide which of the labeled shapes we should use as the source to propagate labels to the target shapes.

\myparagraph{Selection Criteria.}
Given a target $T$, We manually define 4 possible source selection criteria:
\begin{itemize}
    \item \textbf{Nearest Neighbor}: The source shape $S$ that minimizes the Chamfer distance between $S$ and $T$ is selected.
    \item \textbf{Deformation Distance}: The source shape $S$ that minimizes the Chamfer distance between $f_{S,T}(S)$ and $T$ is selected.
    \item \textbf{Cosine Distance}: The source shape $S$ that minimizes the cosine distance distance between the PointNet encodings $v_{S}$ and $v_{T}$ is selected.
    \item \textbf{Cycle Consistency}: The source shape $S$ that minimizes 2-cycle loss for the pair $(S,T)$ is selected.
\end{itemize}
Having selected a pair $(S,T)$, labels can be transferred directly with  our approach. 

\myparagraph{Voting strategy.} Instead of selecting a single source shape to get labels from, combining several voting shapes allows for better segmentation. We select the K-best sources, and make each source shape vote with equal weight for the label of each target point. We evaluate the benefits of this voting approach in Section~\ref{sec:supervisedResults}.

\begin{figure*}[!t!]
\vspace{-1.0cm}
\centering
\captionsetup[subfigure]{justification=centering}
\begin{subfigure}{0.145\linewidth}
\includegraphics[width=\linewidth]{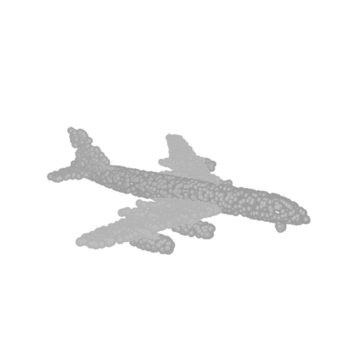} \\
\vspace{-1.8cm}
\includegraphics[width=\linewidth]{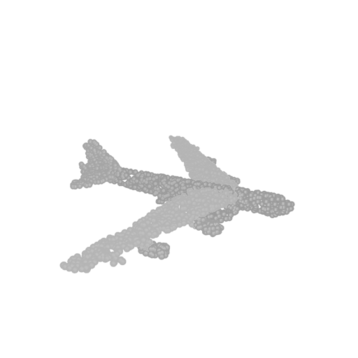} \\
\vspace{-1.2cm}
\includegraphics[width=\linewidth]{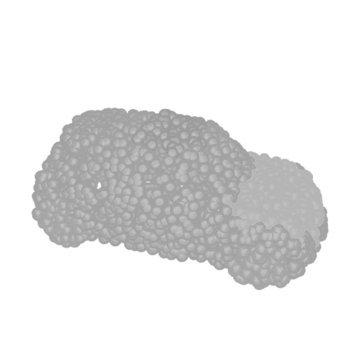} \\
\vspace{-1.6cm}
\includegraphics[width=\linewidth]{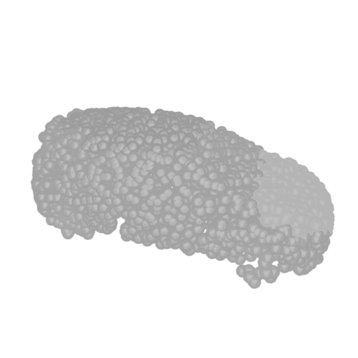} \\
\vspace{-1.0cm}
\includegraphics[width=\linewidth]{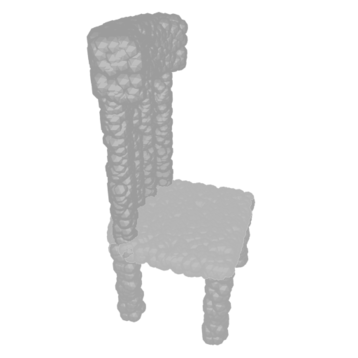} \\
\vspace{-1.0cm}
\includegraphics[width=\linewidth]{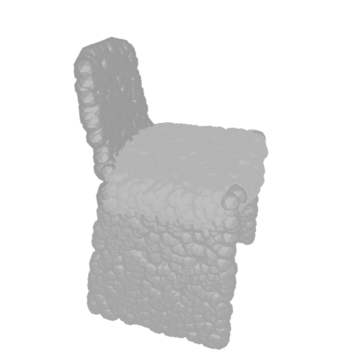} \\
\vspace{-0.4cm}
\includegraphics[width=\linewidth]{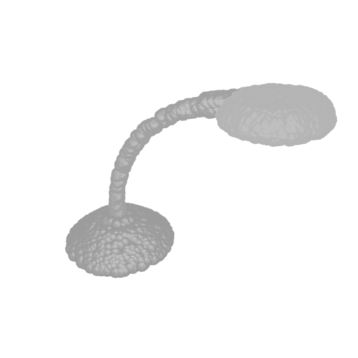} \\
\vspace{-0.8cm}
\includegraphics[width=\linewidth]{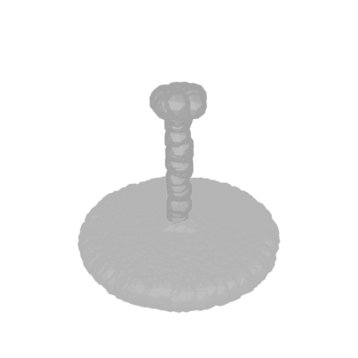} \\
\vspace{-0.8cm}
\includegraphics[width=\linewidth]{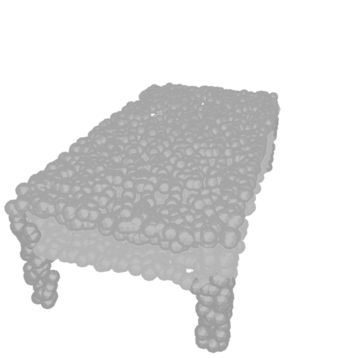} \\
\vspace{-0.5cm}
\includegraphics[width=\linewidth]{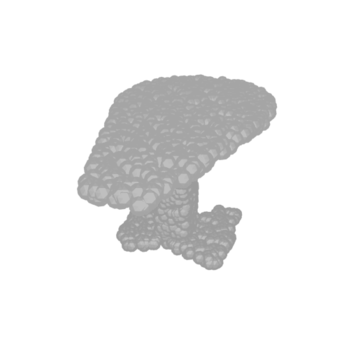} \\
\vspace{-0.7cm}
\caption{Input shape \\ (target)}
\end{subfigure}
\begin{subfigure}{0.145\linewidth}
\includegraphics[width=\linewidth]{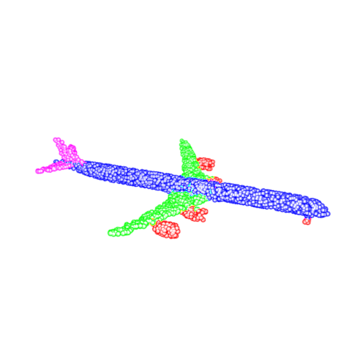} \\
\vspace{-1.8cm}
\includegraphics[width=\linewidth]{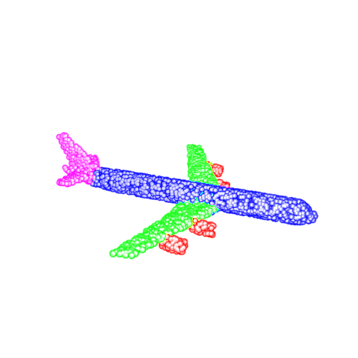} \\
\vspace{-1.2cm}
\includegraphics[width=\linewidth]{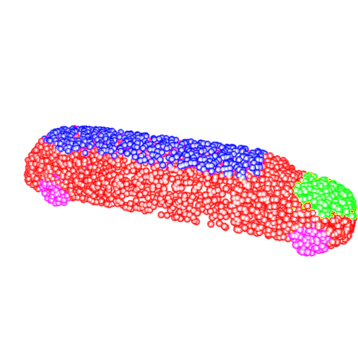} \\
\vspace{-1.6cm}
\includegraphics[width=\linewidth]{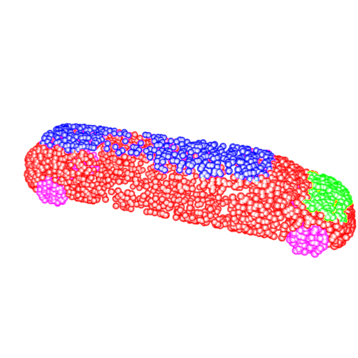} \\
\vspace{-1.0cm}
\includegraphics[width=\linewidth]{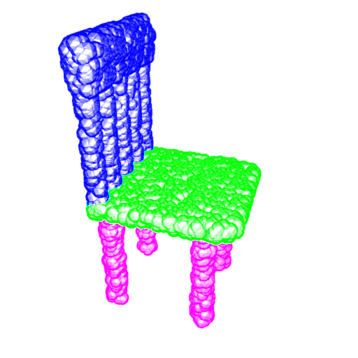} \\
\vspace{-1.0cm}
\includegraphics[width=\linewidth]{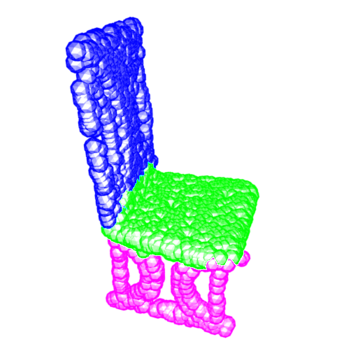} \\
\vspace{-0.4cm}
\includegraphics[width=\linewidth]{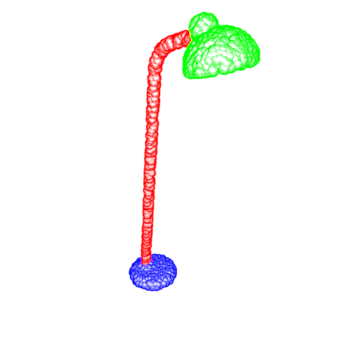} \\
\vspace{-0.8cm}
\includegraphics[width=\linewidth]{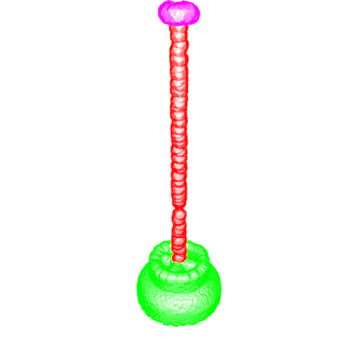} \\
\vspace{-0.8cm}
\includegraphics[width=\linewidth]{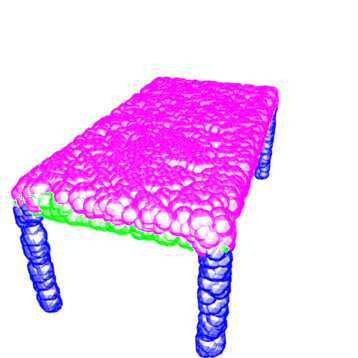} \\
\vspace{-0.5cm}
\includegraphics[width=\linewidth]{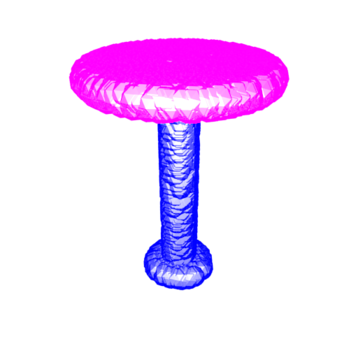} \\
\vspace{-0.7cm}
\caption{Retrieved shape \\ (source)}
\end{subfigure}
\begin{subfigure}{0.145\linewidth}
\includegraphics[width=\linewidth]{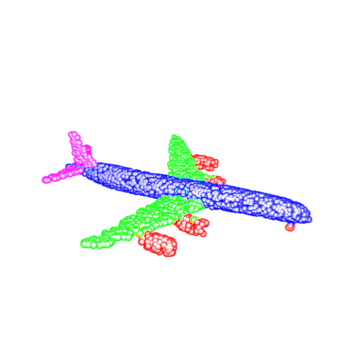} \\
\vspace{-1.8cm}
\includegraphics[width=\linewidth]{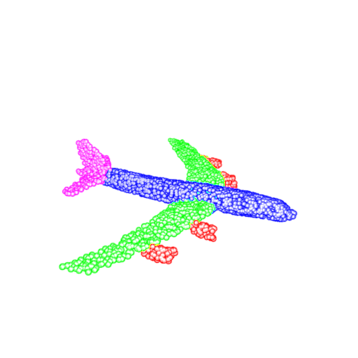} \\
\vspace{-1.2cm}
\includegraphics[width=\linewidth]{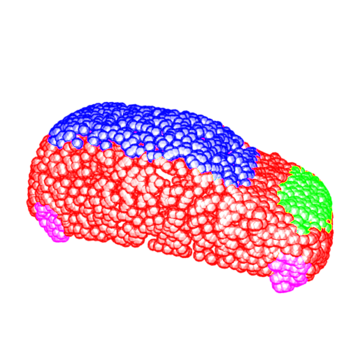} \\
\vspace{-1.6cm}
\includegraphics[width=\linewidth]{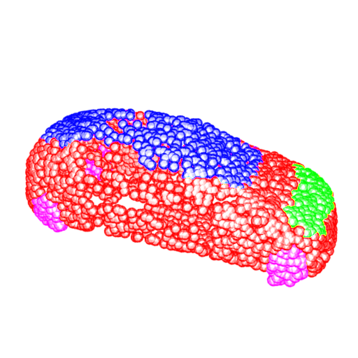} \\
\vspace{-1.0cm}
\includegraphics[width=\linewidth]{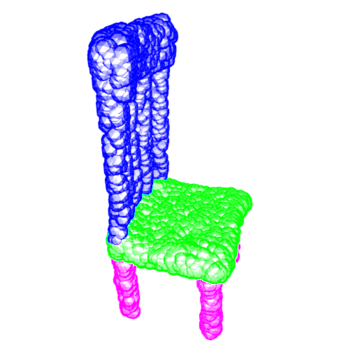} \\
\vspace{-1.0cm}
\includegraphics[width=\linewidth]{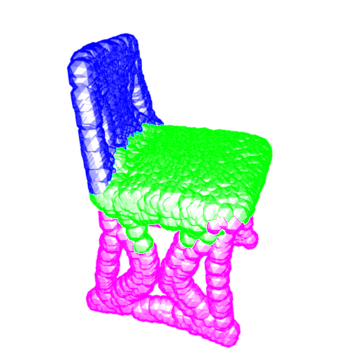} \\
\vspace{-0.4cm}
\includegraphics[width=\linewidth]{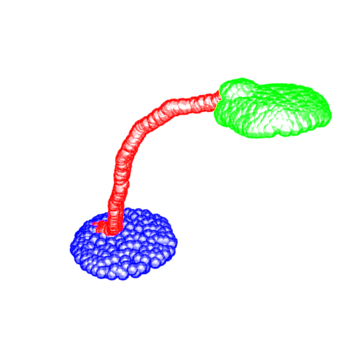} \\
\vspace{-0.8cm}
\includegraphics[width=\linewidth]{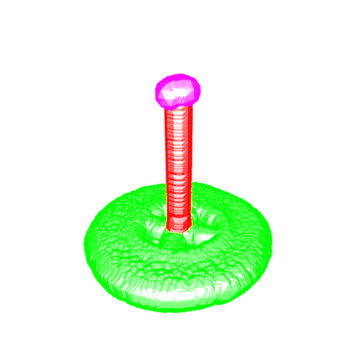} \\
\vspace{-0.8cm}
\includegraphics[width=\linewidth]{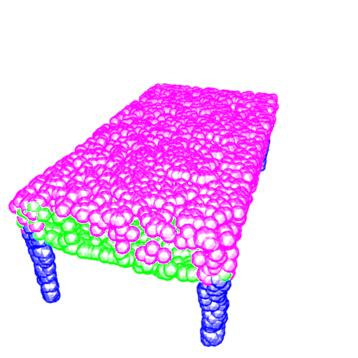} \\
\vspace{-0.5cm}
\includegraphics[width=\linewidth]{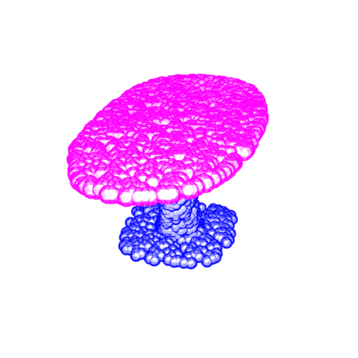} \\
\vspace{-0.7cm}
\caption{Deformed \\ retrieved shape}
\end{subfigure}
\begin{subfigure}{0.145\linewidth}
\includegraphics[width=\linewidth]{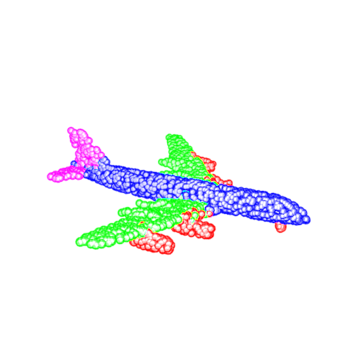} \\
\vspace{-1.8cm}
\includegraphics[width=\linewidth]{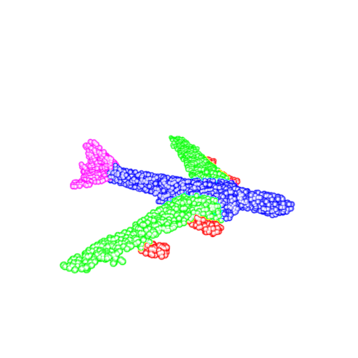} \\
\vspace{-1.2cm}
\includegraphics[width=\linewidth]{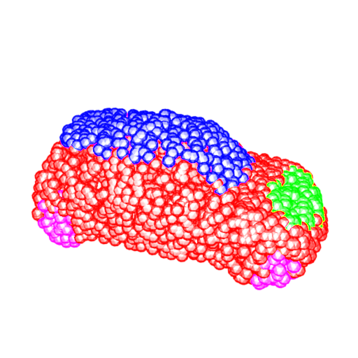} \\
\vspace{-1.6cm}
\includegraphics[width=\linewidth]{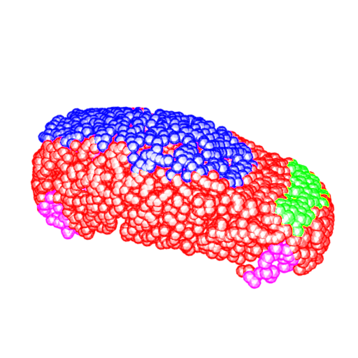} \\
\vspace{-1.0cm}
\includegraphics[width=\linewidth]{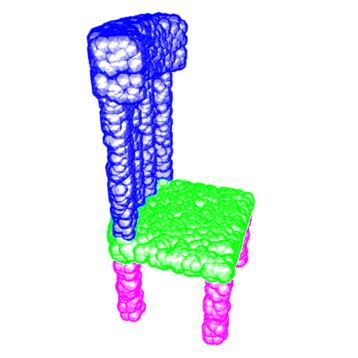} \\
\vspace{-1.0cm}
\includegraphics[width=\linewidth]{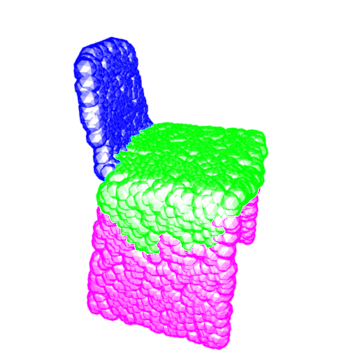} \\
\vspace{-0.4cm}
\includegraphics[width=\linewidth]{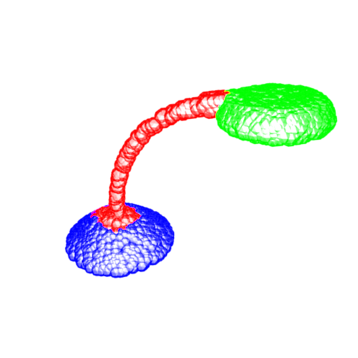} \\
\vspace{-0.8cm}
\includegraphics[width=\linewidth]{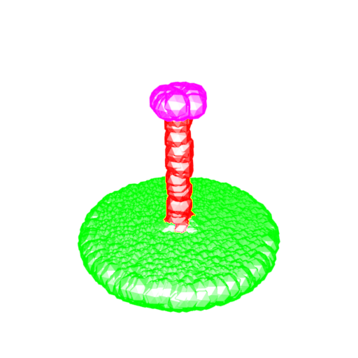} \\
\vspace{-0.8cm}
\includegraphics[width=\linewidth]{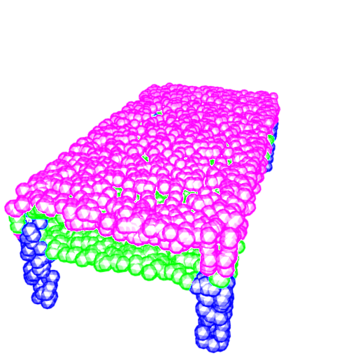} \\
\vspace{-0.5cm}
\includegraphics[width=\linewidth]{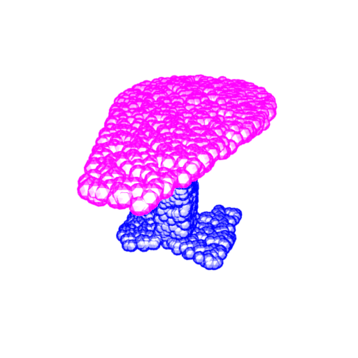} \\
\vspace{-0.7cm}
\caption{Transferred labels \\ to input}
\end{subfigure}
\begin{subfigure}{0.145\linewidth}
\includegraphics[width=\linewidth]{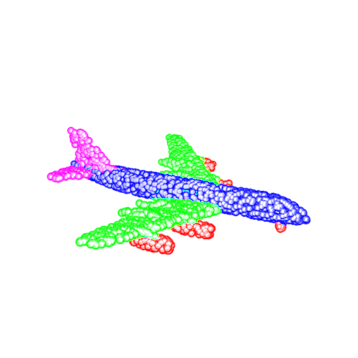} \\
\vspace{-1.8cm}
\includegraphics[width=\linewidth]{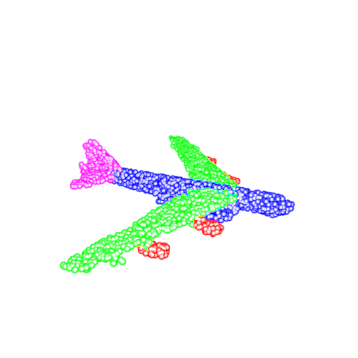} \\
\vspace{-1.2cm}
\includegraphics[width=\linewidth]{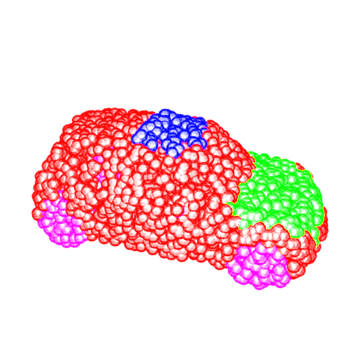} \\
\vspace{-1.6cm}
\includegraphics[width=\linewidth]{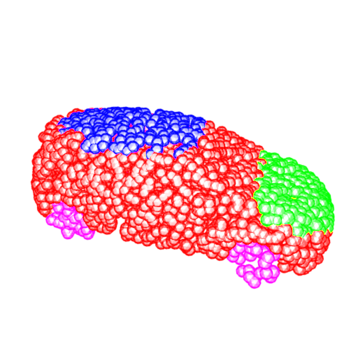} \\
\vspace{-1.0cm}
\includegraphics[width=\linewidth]{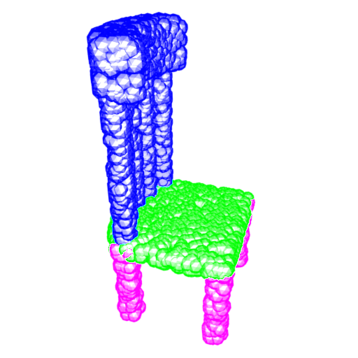} \\
\vspace{-1.0cm}
\includegraphics[width=\linewidth]{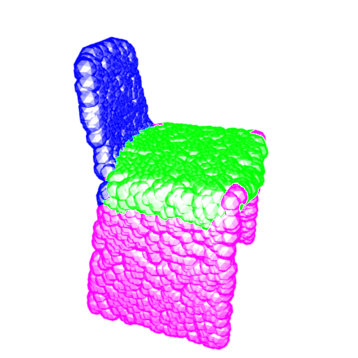} \\
\vspace{-0.4cm}
\includegraphics[width=\linewidth]{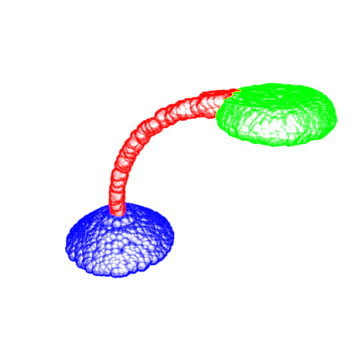} \\
\vspace{-0.8cm}
\includegraphics[width=\linewidth]{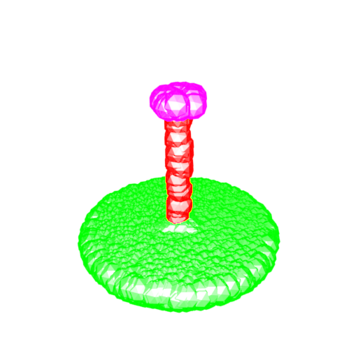} \\
\vspace{-0.8cm}
\includegraphics[width=\linewidth]{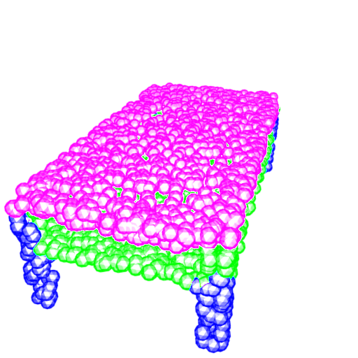} \\
\vspace{-0.5cm}
\includegraphics[width=\linewidth]{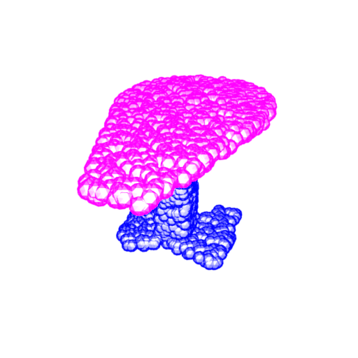} \\
\vspace{-0.7cm}
\caption{Ground truth \\ labels}
\end{subfigure}
\begin{subfigure}{0.145\linewidth}
\includegraphics[width=\linewidth]{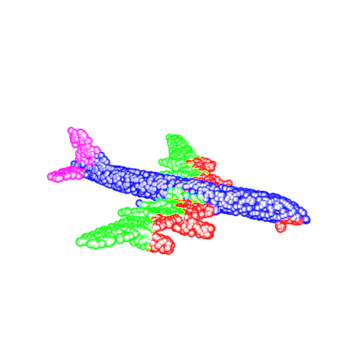} \\
\vspace{-1.8cm}
\includegraphics[width=\linewidth]{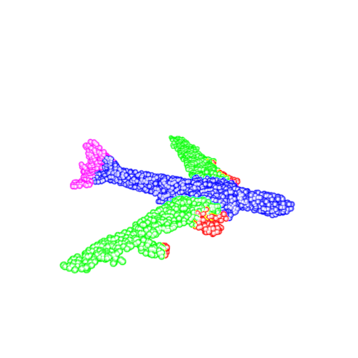} \\
\vspace{-1.2cm}
\includegraphics[width=\linewidth]{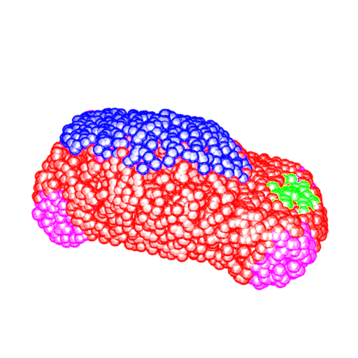} \\
\vspace{-1.6cm}
\includegraphics[width=\linewidth]{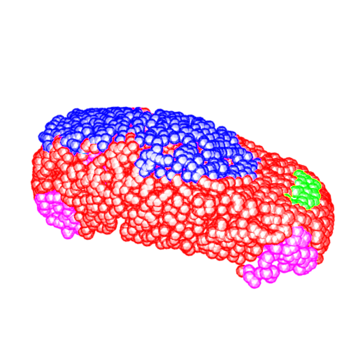} \\
\vspace{-1.0cm}
\includegraphics[width=\linewidth]{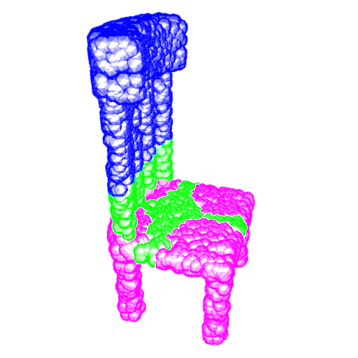} \\
\vspace{-1.0cm}
\includegraphics[width=\linewidth]{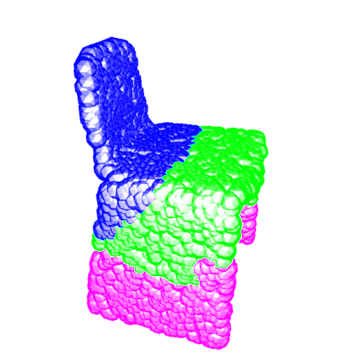} \\
\vspace{-0.4cm}
\includegraphics[width=\linewidth]{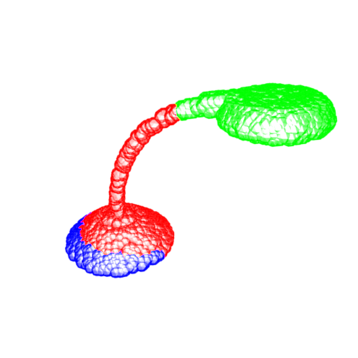} \\
\vspace{-0.8cm}
\includegraphics[width=\linewidth]{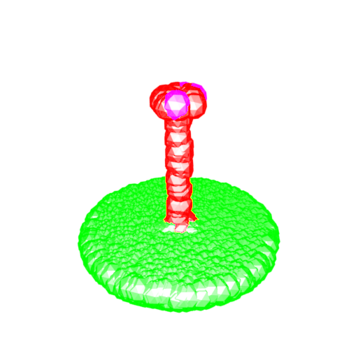} \\
\vspace{-0.8cm}
\includegraphics[width=\linewidth]{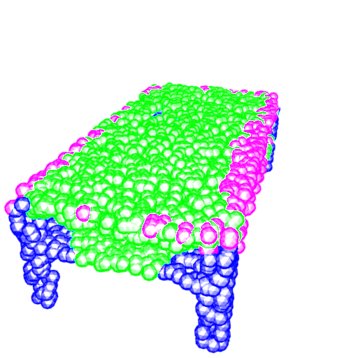} \\
\vspace{-0.5cm}
\includegraphics[width=\linewidth]{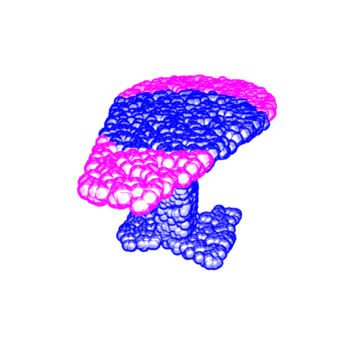} \\
\vspace{-0.7cm}
\caption{Identity baseline \\ segmentation}
\end{subfigure}
\caption{\textbf{Qualitative results.} 
For each input shape (a), we select the top nearest neighbor from 400 training examples with part segmentations using the cycle-consistency criterion (b). We apply our approach to deform the retrieved shape to align with the input shape (c). Given the deformed shape, we transfer the labels onto the input shape (d). For each category, we show the top results that maximize IoU with the ground truth (e). For comparison, we show the Identity baseline in (f). Notice how our method successfully transfers labels and improves over the baseline.
}
 \label{fig:bestresultsoveridentity}
 \vspace*{-4mm}
\end{figure*}

\begin{figure*}[!t!]
\centering
\captionsetup[subfigure]{justification=centering}
\begin{subfigure}{0.15\linewidth}
\includegraphics[width=\linewidth]{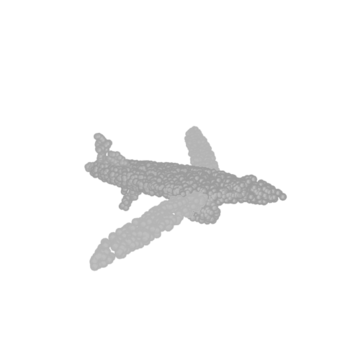} \\
\vspace{-1.2cm}
\includegraphics[width=\linewidth]{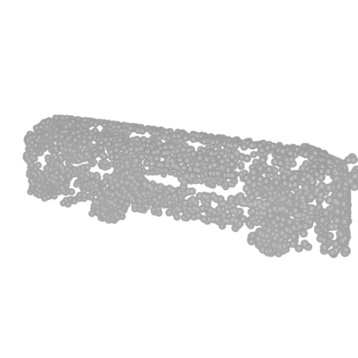} \\
\vspace{-1.0cm}
\includegraphics[width=\linewidth]{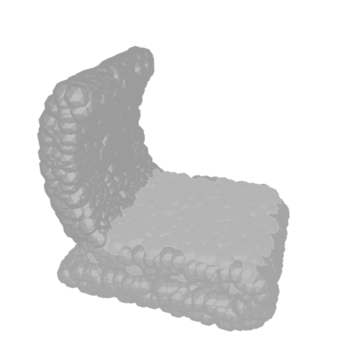} \\
\vspace{-0.4cm}
\includegraphics[width=\linewidth]{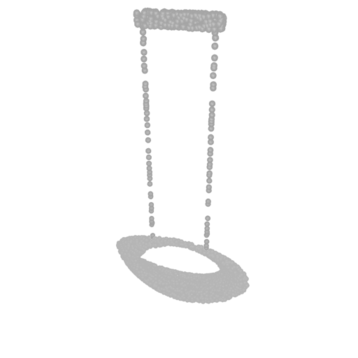} \\
\vspace{-0.8cm}
\includegraphics[width=\linewidth]{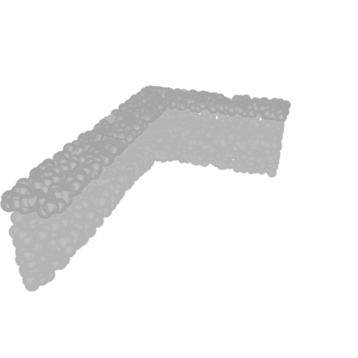} \\
\vspace{-0.7cm}
\caption{Input shape \\ (target)}
\end{subfigure}
\begin{subfigure}{0.15\linewidth}
\includegraphics[width=\linewidth]{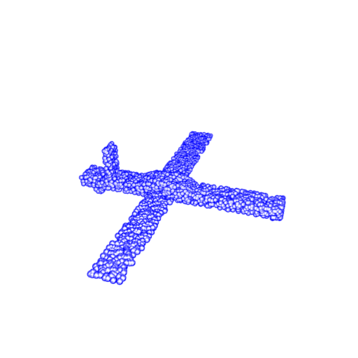} \\
\vspace{-1.2cm}
\includegraphics[width=\linewidth]{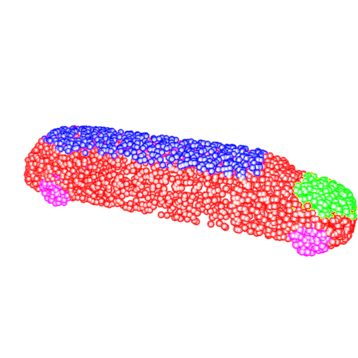} \\
\vspace{-1.0cm}
\includegraphics[width=\linewidth]{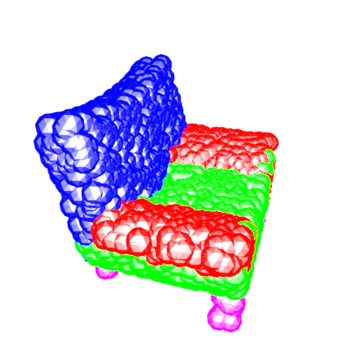} \\
\vspace{-0.4cm}
\includegraphics[width=\linewidth]{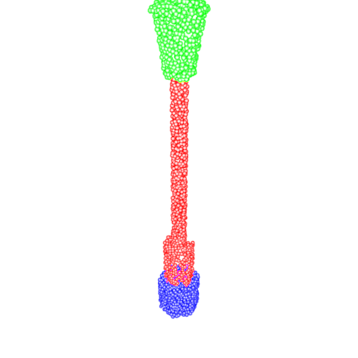} \\
\vspace{-0.8cm}
\includegraphics[width=\linewidth]{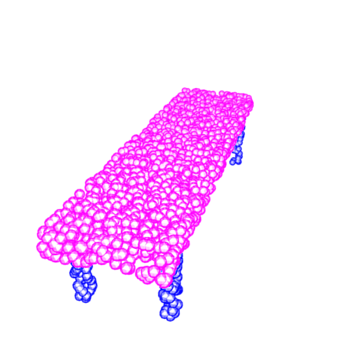} \\
\vspace{-0.7cm}
\caption{Retrieved shape \\ (source)}
\end{subfigure}
\begin{subfigure}{0.15\linewidth}
\includegraphics[width=\linewidth]{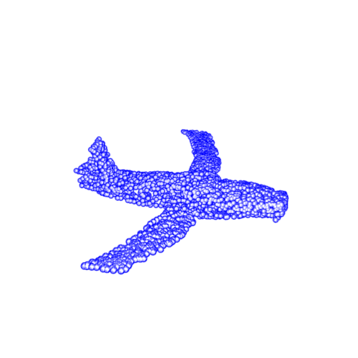} \\
\vspace{-1.2cm}
\includegraphics[width=\linewidth]{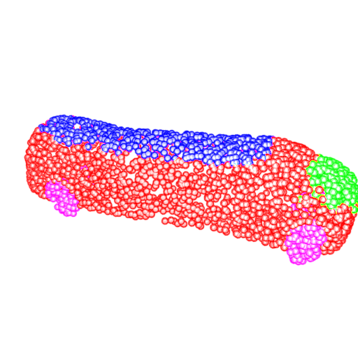} \\
\vspace{-1.0cm}
\includegraphics[width=\linewidth]{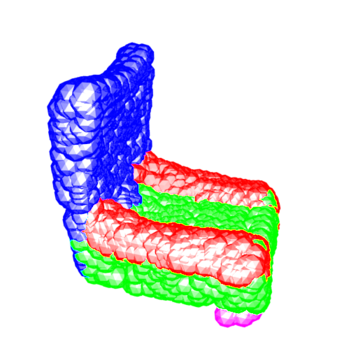} \\
\vspace{-0.4cm}
\includegraphics[width=\linewidth]{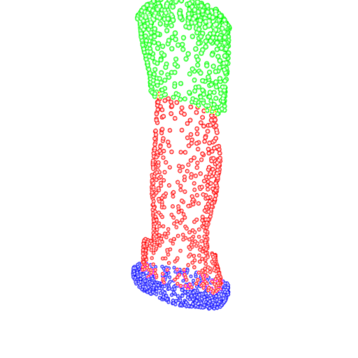} \\
\vspace{-0.8cm}
\includegraphics[width=\linewidth]{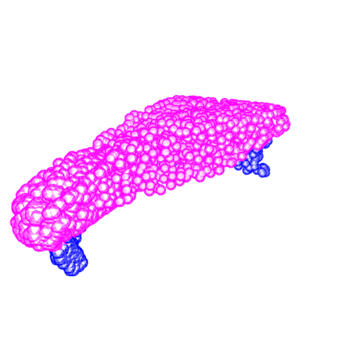} \\
\vspace{-0.7cm}
\caption{Deformed \\ retrieved shape}
\end{subfigure}
\begin{subfigure}{0.15\linewidth}
\includegraphics[width=\linewidth]{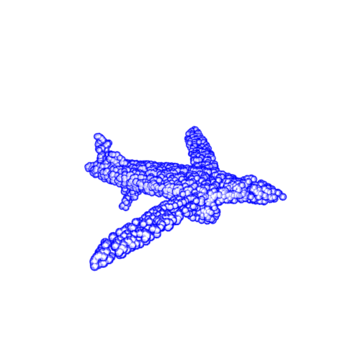} \\
\vspace{-1.2cm}
\includegraphics[width=\linewidth]{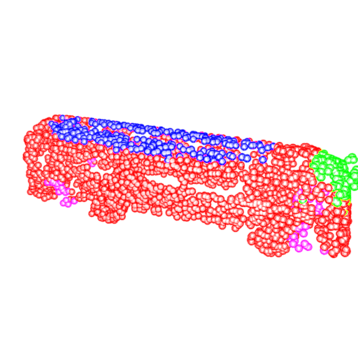} \\
\vspace{-1.0cm}
\includegraphics[width=\linewidth]{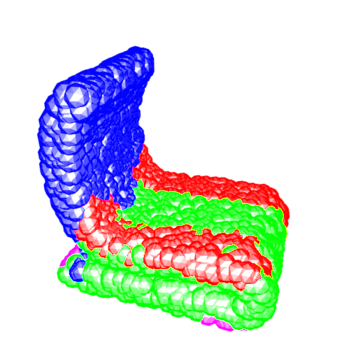} \\
\vspace{-0.4cm}
\includegraphics[width=\linewidth]{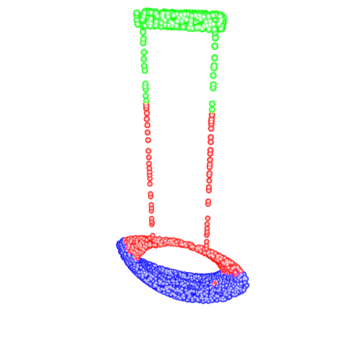} \\
\vspace{-0.8cm}
\includegraphics[width=\linewidth]{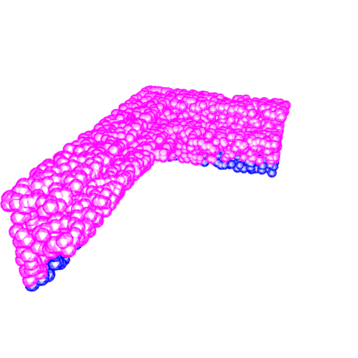} \\
\vspace{-0.7cm}
\caption{Transferred labels \\ to input}
\end{subfigure}
\begin{subfigure}{0.15\linewidth}
\includegraphics[width=\linewidth]{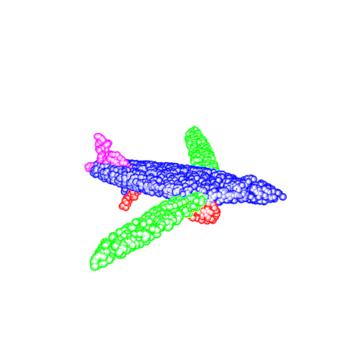} \\
\vspace{-1.2cm}
\includegraphics[width=\linewidth]{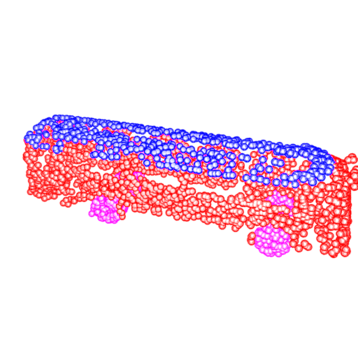} \\
\vspace{-1.0cm}
\includegraphics[width=\linewidth]{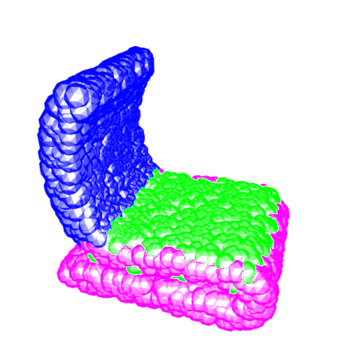} \\
\vspace{-0.4cm}
\includegraphics[width=\linewidth]{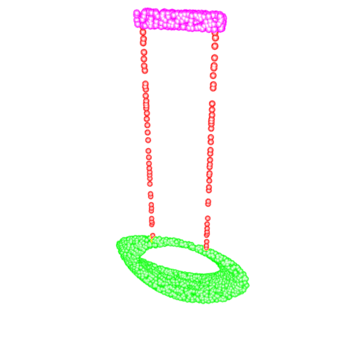} \\
\vspace{-0.8cm}
\includegraphics[width=\linewidth]{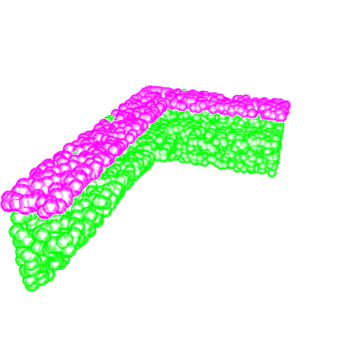} \\
\vspace{-0.7cm}
\caption{Ground truth \\ labels}
\end{subfigure}
\begin{subfigure}{0.15\linewidth}
\includegraphics[width=\linewidth]{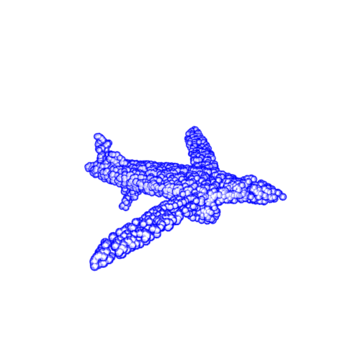} \\
\vspace{-1.2cm}
\includegraphics[width=\linewidth]{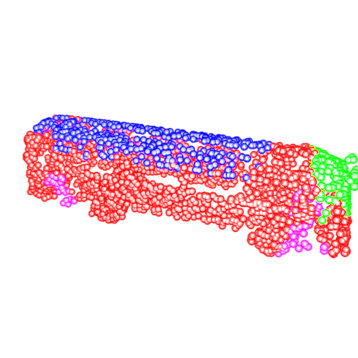} \\
\vspace{-1.0cm}
\includegraphics[width=\linewidth]{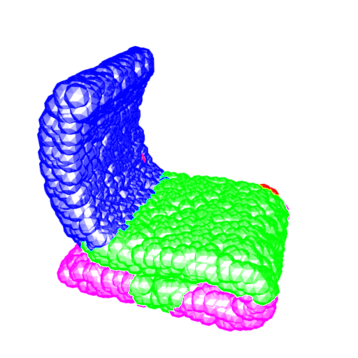} \\
\vspace{-0.4cm}
\includegraphics[width=\linewidth]{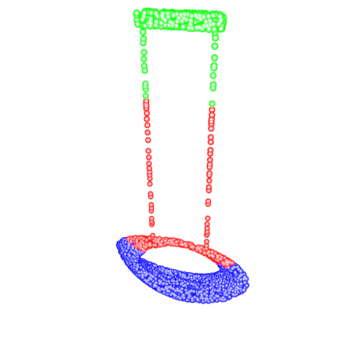} \\
\vspace{-0.8cm}
\includegraphics[width=\linewidth]{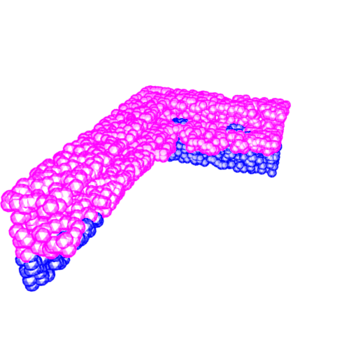} \\
\vspace{-0.7cm}
\caption{Identity baseline \\ segmentation}
\end{subfigure}
\caption{\textbf{Failures.} 
Example failures include when a retrieved shape has inconsistent annotation (rows 1,2,5) and poor deformation due to different topology (rows 3,4).
}
 \label{fig:failurecases}
 \vspace*{-4mm}
\end{figure*}

\section{Results}

In this section, we show qualitative and quantitative results on the tasks of few-shot and supervised semantic segmentation and compare against several baselines. 

\myparagraph{Data and evaluation criteria.} We evaluated our approach on the standard ShapeNet part dataset~\cite{Yi16}. We restricted ourselves to the 5 most populated categories, namely Airplane, Car, Chair, Lamp, and Table. Point clouds sampled on mesh objects are densely labeled for segmentation with one to five parts. 
We follow Qi et al.~\cite{qi2016pointnet} and report the mean intersection over union (mIoU) between the predicted and ground truth segmentation across instances in a category.

\myparagraph{Baselines.}  We compare our unsupervised approach against supervised and unsupervised approaches. %
We used PointNet as a supervised baseline. Our unsupervised baselines include a learned approach derived from Atlasnet~\cite{groueix2018} and variants of iterative closest points (ICP)~\cite{Zhang94iterativepoint, Besl:1992:MRS:132013.132022}. AtlasNet is a template-based reconstruction method that predicts a transformation of the template matching the target shape. The learned deformations have been previously observed to be semantically consistent~\cite{groueix2018b}. To transfer segmentation labels from a source to a target, we project the source labels on the source reconstruction through nearest neighbors, then on the template through dense correspondence between the template and the source reconstruction. Similarly, we transfer labels on the template to the target by dense correspondence and nearest neighbors. AtlasNet is trained on the same train/test splits as our approach. We consider two settings of AtlasNet -- with 10 patches or 1 sphere as the template.
Additionally, we use two standard shape alignment baselines. First, labels can be transferred from source to target through nearest neighbor matching, which we call the {\em Identity} baseline. An immediate refinement over this baseline is to apply ICP to align the source to the target, and then use nearest neighbors. We call the latter the {\em ICP} baseline.

\subsection{Qualitative Results}

\begin{figure}[!h]
\centering
\includegraphics[width=\linewidth]{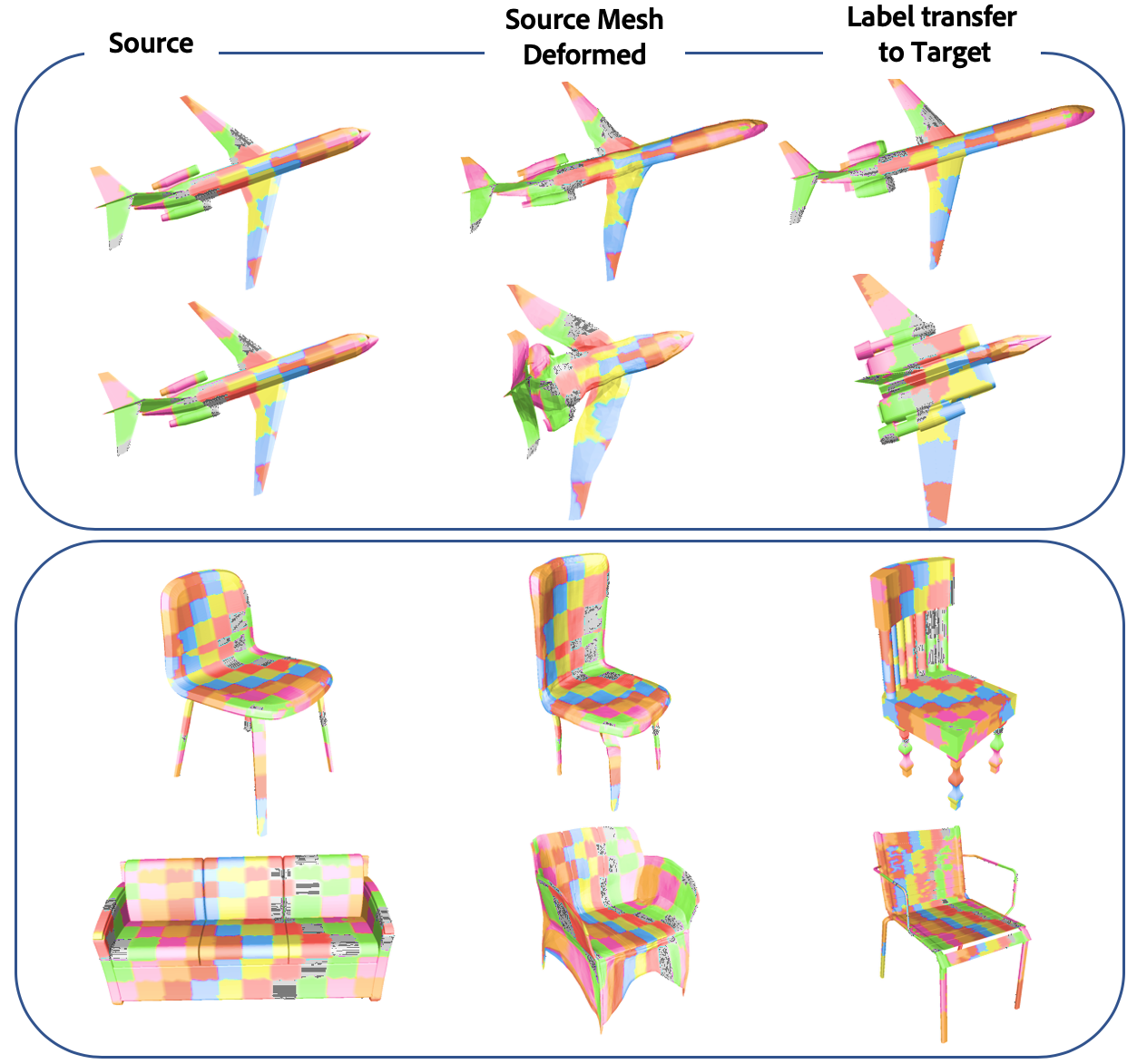} \\
\caption{\final{\textbf{Mapping function quality.} We apply a checkerboard colorization scheme on the source (\textbf{left}), and use our approach to deform (\textbf{middle}) the source shape  to the target shape (\textbf{right}). The labels are transferred from the deformed shape to the target shape through nearest neighbors. For each category, we show a example of good reconstruction (\textbf{top}) and poor reconstruction (\textbf{bottom}). Notice the high quality of the mapping in both cases.\\ 
}}
 \label{fig:high_freq}
\end{figure}
\begin{figure}[!h]
\centering
\includegraphics[width=\linewidth]{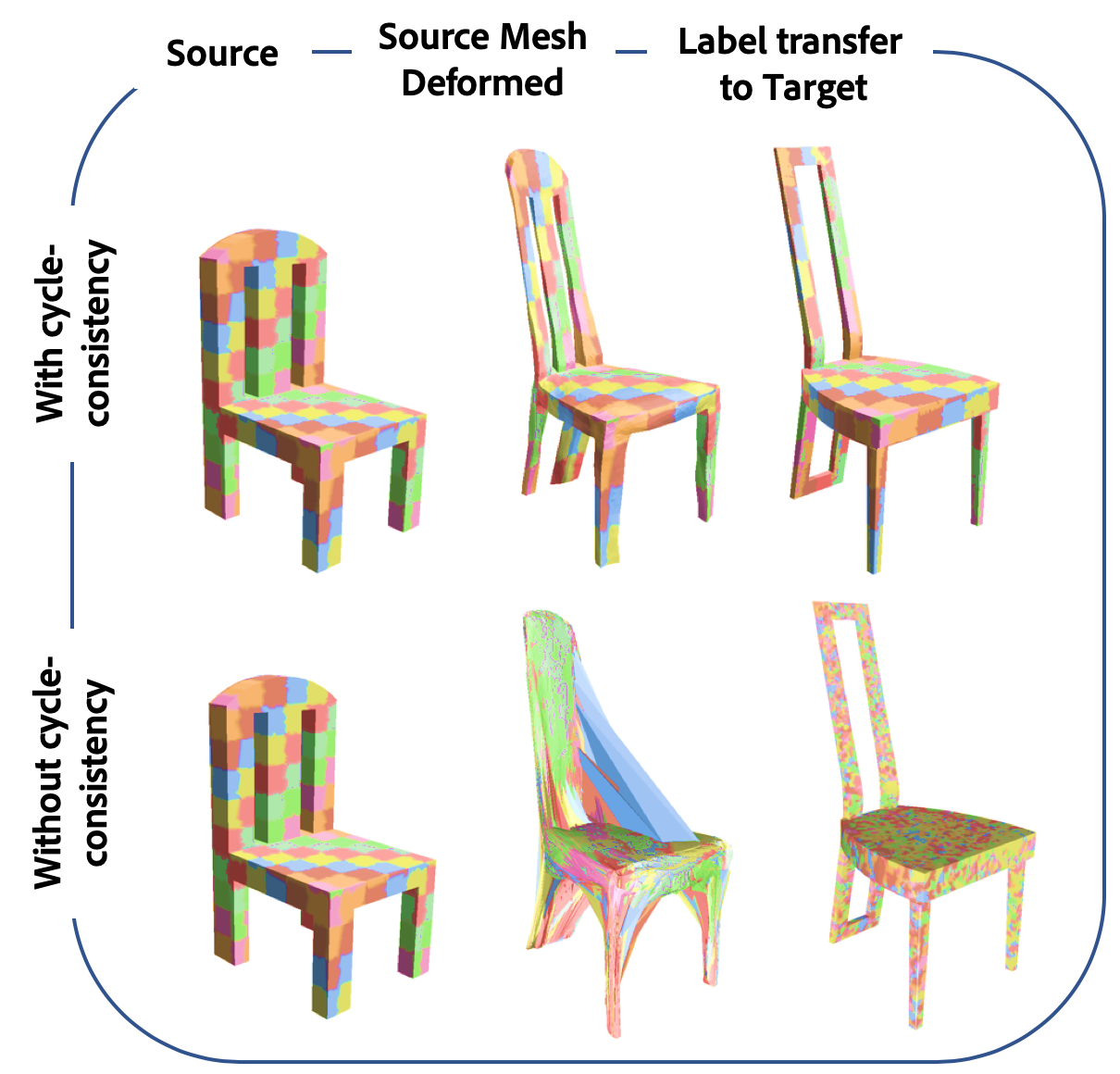} \\
\caption{\final{\textbf{Cycle-consistency performance.} We apply a checkerboard colorization scheme on the source (\textbf{left}), and use our approach with cycle-constistency (\textbf{top}) and without (\textbf{bottom}) to deform (\textbf{middle}) the source shape  to the target shape (\textbf{right}). The labels are transferred from the deformed shape to the target shape through nearest neighbors. \\ 
}}
 \label{fig:distorsion}
\end{figure}

\myparagraph{Correspondences.} \final{In figure~\ref{fig:high_freq} we visualize in more detail the correspondences obtained with our approach. We visualize how each point on the source shape is deformed and transferred to the target shape using a colored checkerboard. For each example, we show a successful deformation (top) and a failure case (bottom). Note how the checkerboard appears nicely deformed in the case of successful deformation, and still appears consistent on some parts in the failure cases. 
}

\myparagraph{Cycle-consistency.} \final{In figure~\ref{fig:distorsion} we compare the mappings learned by our approach with and without cycle-consistency loss. The Chamfer Distance is a point based loss with no control over the amount of distorsion. Notice in this case that the deformed source has large triangles. It indicates that the mapping learned by a Chamfer loss alone is not smooth, and can't be used in label tranfer. On the other hand, the cycle-consistency loss leads to a smooth and high quality mapping.}\\

\myparagraph{Segmentation transfer.} When looking at the results, a first surprising observation is the high quality of the identity baseline (this is quantitatively confirmed in Table~\ref{tab:supervisedsegmentation}). Indeed, the different criteria tend to select shapes that are really close to the target.  To focus on interesting examples, we selected in Figure~\ref{fig:bestresultsoveridentity} the pairs that maximize the performance improvement provided by our method compared to the identity baseline using the cycle-consistency-selection criterion. The richness of the learned deformations allows our method to find meaningful correspondences in cases where the training example is far from the target shape and the identity baseline does not work. Note that the deformations are often far from isometric. Thus, methods that rely on regularization toward identity, a popular approach to regularize learned deformations~\cite{groueix2018b,cmrKanazawa18,wang2018pixel2mesh}, would likely fail. %

\myparagraph{Failure cases.} Figure~\ref{fig:failurecases} shows failures of our method. We show for each category the pair $(S,T)$ which minimizes our segmentation transfer performance. It is clear that the corresponding shapes are rare and specific object instances. \final{We observe two main sources of errors. First, in some cases where we correctly deform $S$ in $T$, the ground truth labeling was inconsistent, leading to large errors. For example, notice how the source airplane has a single label. Second, $S$ and $T$ are sometimes too distant topologically so that a high-fidelity reconstruction of $T$ is impossible by deforming $S$. For example, notice how the pole of the lamp has been erroneously inflated to match the target shape.}

\subsection{Quantitative Results}

\subsubsection{Few-shot Segmentation}
In this section, we evaluate our approach on the task of transferring semantic labels from a small set of segmented shapes to unlabeled data.

\begin{table*}[!ht]
\vskip 0.1in
\begin{center}
\begin{small}
\begin{tabular}{lccccccr}
\toprule
 10 shots & Selection Criterion & Airplane & Car & Chair & Lamp & Table \\
\midrule
(a) Pointnet  & - & $14.0\pm 8.0$ & $11.7 \pm 10.4$ & $21.1 \pm  13.1$ & $26.0 \pm 13.2$ & $43.5 \pm 15.5$\\
\midrule
(b) Atlasnet Patch  & Nearest Neighbors & $62.6 \pm 2.4$ &  $52.3 \pm 9.1 $  & $72.1  \pm 1.2 $  & $62.8 \pm2.2 $  & $61.6  \pm 3.7 $  & \\
(c) Atlasnet Sphere  & Nearest Neighbors &$62.2 \pm 2.2$  & $ 52.9 \pm9.1$ & $ 70.2  \pm 1.2$  & $ 59.3 \pm1.8$  & $ 60.0  \pm 5.1$  &   \\
(d) ICP & Nearest Neighbors &  $65.5  \pm 3.1 $ & $61.3  \pm  1.1 $ & $75.8 \pm 1.2 $ & $64.8  \pm  5.0 $ & $64.9 \pm 3.9 $ &   \\
(e) Ours  & Nearest Neighbors  &  $ \bm{67.1  \pm  2.9} $ & $ \bm{61.4  \pm  1.1} $ & $ \bm{78.9  \pm  1.1} $ & $ \bm{65.8 \pm 5.2} $ & $ \bm{66.1 \pm 4.5} $ &  \\
\midrule
(f)  Ours & Cycle Consistency & $ 67.9 \pm 3.0 $ & $60.2 \pm 3.4$  &$ 81.8 \pm 0.7 $ & $ 69.1 \pm 5.4 $ & $ 68.8 \pm 4.0 $ &  \\
(g) Ours & Oracle & $ 74.9 \pm 3.0$  & $68.6 \pm 2.4$ & $ 86.4 \pm 0.6 $ & $ 80.3 \pm 3.8 $ & $ 77.8 \pm 2.1 $ &  \\
\midrule
\bottomrule
\end{tabular}
\end{small}
\end{center}
\caption{\textbf{Few-shot segmentation:}. We compare \textbf{(e, f)} our approach with \textbf{(a)} Pointnet~\cite{qi2016pointnet}, a supervised method, trained per category, \textbf{(b, c)} two unsupervised baselines based on Atlasnet \cite{groueix2018} and \textbf{(e)} ICP. We pre-train all \textbf{(b, c , e, f)} unsupervised approaches on the train splits (without labels). Given a target shape $T$ and 10 segmented train samples, we select $T$'s nearest neighbors S. In Atlasnet \textbf{(b, c)}, labels are propagated through the template. In our approach \textbf{(e, f, g)}, labels are propagated  from $T_S$ to T. We report in (\textbf{g}) the best performance of our method over the 10 shots. The mean IoU is reported. Results are averaged over 10 runs.}
 \label{tab:fewshot}
\end{table*}

We report quantitative results for few-shot semantic segmentation on point clouds in Table~\ref{tab:fewshot}. Note that the learning-based methods are all trained separately for each category. Since the results depend on the sampled shapes used in the training set, we report the average and standard deviation over ten randomly sampled training sets. We use the Nearest Neighbors criterion to pair sources and targets and compare our approach against all baselines \textbf{(b, c, d, e)}. Notice that our approach out-performs all baselines on all categories. Interestingly, the AtlasNet baseline is not on par with ICP, hinting at the difficulty of predicting two consistent deformations of the template. 

We find that the Cycle Consistency criterion \textbf{(f)} is a stronger selection criterion than Nearest Neighbors and boosts the results simply by selecting a better (Source, Target) pair. We also report an oracle source-shape selection with our approach where the source shape maximising IoU with the target is selected, which corresponds to the scenario where an optimal source shape is selected. Notice the large improvement of the oracle, showing the quality of our deformations and the potential of our method.

\begin{figure*}[ht!!]
\centering
 \includegraphics[width=0.30\linewidth]{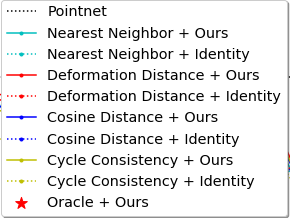}
 \includegraphics[width=0.3\linewidth]{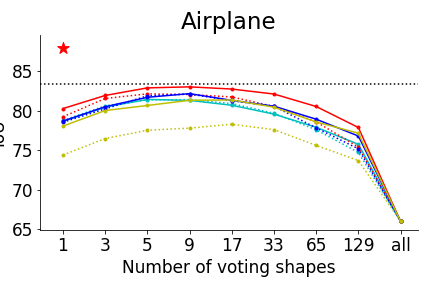}
 \includegraphics[width=0.3\linewidth]{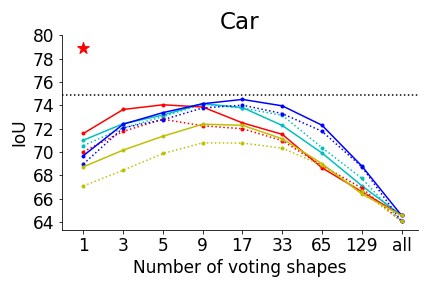}\\
 \includegraphics[width=0.3\linewidth]{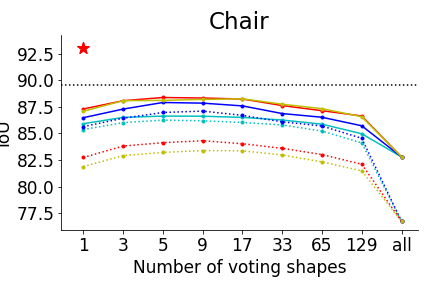}
 \includegraphics[width=0.3\linewidth]{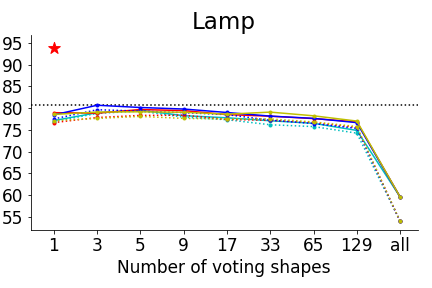}
 \includegraphics[width=0.3\linewidth]{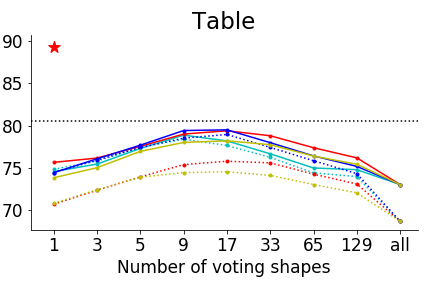}
\caption{\textbf{Criteria and voting strategies.} Study of the number of voting shapes for the transfer of segmentation label, across 4 criteria (see~\ref{sec:segmentation}) - Nearest Neighbors, Deformation Distance, Cosine Distance and Cycle Consistency -, and across 5 Shapenet categories. Our transformation method (solid lines) almost always enhance the identity baseline (dashed lines). We report a supervised baseline, Pointnet~\cite{qi2016pointnet} and the oracle source which maximizes IoU for our method. Notice how the oracle significantly outperforms the Pointnet baseline, making the search of a strong selection criterion a good direction. Our models are category specific and trained without segmentation supervision. All of the train set is searched to maximize each criterion.}
  \label{fig:ensemble}
  \vspace*{-4mm}
\end{figure*}

\subsubsection{Supervised segmentation}
\label{sec:supervisedResults}

Our method is not designed to be competitive when many training samples are available. Indeed, it solves for the deformation against each of the provided segmented shapes, which for large numbers of examples can be computationally expensive compared to feed-forward segmentation predictions like PointNet~\cite{qi2016pointnet}. \final{One forward pass through our network deforms a source shape in a target shape in 7 milliseconds (ms), with a
7ms standard deviation (std). ICP takes 28 ms with a 17 std\footnote{
We use Open3D~\cite{Zhou2018} to compute ICP ran on Intel i7-6900K - 3.2 GHz and run our method on an NVIDIA TITAN X.}.} Here, however, we study the performance of our method in this case, using the segmentation of the many training shapes as supervision during training and making the ten best shapes vote during testing. 
We report results of our unsupervised method. In addition, we consider adding supervision to our approach by computing Chamfer distances over points with the same segmentation label.  
The corresponding results are reported in Table~\ref{tab:supervisedsegmentation}
Table~\ref{tab:supervisedsegmentation} shows that, when using all the annotations, nearest neighbors is again a surprisingly good baseline, only slightly below performance of PointNet. %
Despite the good performance of the identity baseline, our method outperforms it in all categories and performs on par with PointNet. Note that the encoders of our approach incorporate two PointNet architectures, which makes this result intuitive.

Table~\ref{tab:supervisedsegmentation} also highlights the importance of the criterion selection. Notice the significant boost in each category gained by carefully choosing the selection criterion over the Nearest Neighbors criterion. The exciting performance of the oracle, way over the PointNet baseline, is another incentive at carefully designing selection criteria.

Finally, notice that our unsupervised trained model is on par with our supervised one. The boost gained by supervised training is marginal except in the car category. It confirms that our cycle-consistent loss is efficient to enforce meaningful part correspondence.

\begin{table}[!h]
\vskip 0.1in
\begin{center}
\resizebox{\columnwidth}{!}{
\begin{small}
\begin{tabular}{lccccccr}
\toprule
  & Selection & Airplane & Car & Chair & Lamp & Table \\
\midrule
(a) Pointnet  & - & 83.4 & 74.9 & \textbf{89.6} & \textbf{80.8} & \textbf{80.6} \\
\midrule
(b) Identity & NN & 81.3 &  74.0 & 86.1 & 78.4 & 78.9 \\
\midrule
(c) Ours unsup & NN& 81.5 &  73.9 & 86.6 & 78.8 & 79.2 \\
(d)  Ours unsup & Best criterion  & 83.4 & 74.6 & 88.4 & 79.8 & 79.7  \\
(e)  Ours unsup & Oracle& 87.9 &  78.9 & 93.0 & 93.9 & 89.3 \\
\midrule
(f)  Ours sup& NN  & 81.2 & 75.9 & 86.9 & 78.4 & 79.0\\
(g) Ours sup & Best criterion & \textbf{83.5} & \textbf{76.4} & 88.8 & 79.3 & 79.9 \\
(h) Ours sup& Oracle &  88.0 & 80.2 & 93.1 & 93.4 & 89.4 \\
\midrule
\bottomrule
\end{tabular}
\end{small}
}
\end{center}
\caption{\textbf{Supervised segmentation:}. We compare our approach with \textbf{(a)} Pointnet~\cite{qi2016pointnet} and \textbf{(b)} Identity baseline. Our approach can be trained with part supervision \textbf{(f, g, h)} or without \textbf{(c, d, e)}. Given a target shape $T$ and the segmented train set, we compare 3 types of source shapes : \textbf{(b, c, f)} $T$'s Nearest Neighbors; \textbf{(d, g)} the best shape among all criteria see~\ref{sec:segmentation}; and \textbf{(e, h)}  the \textit{a posteriori} best shape over all train sample. A voting strategy is used on the top 10 shapes in \textbf{(b, c, d, f, g)}. The mean IoU is reported.}
\label{tab:supervisedsegmentation}
\end{table}

\subsubsection{Selection criteria and voting strategy}
Figure~\ref{fig:ensemble} shows a quantitative comparison on all criteria, on all category for the identity baseline and our approach using a voting strategy with different number of shapes. The oracle, and PointNet performances are also reported. The Deformation Distance criterion outperforms all other criteria but remains far from the oracle. The oracle performs better than the PointNet baseline across all categories. As a sanity check, we observe that our method outperforms the identity baseline in all settings, showing that it helps to apply our method to transfer labels from $S$ to $T$.

Figure~\ref{fig:ensemble} also confirms that using several source shapes is beneficial when many annotated examples are available. In the limit, when all source shapes vote and selection criterion does not matter anymore, an average labelling is predicted with poor performances, which again outlines the importance of source selection. Using nine source shapes performs the best across most criteria and categories when all the training annotations can be used.

\subsection{Ablation Study}
\label{sec:ablation}
In this section we conduct an ablation study to empirically validate our approach. Table~\ref{tab:ablation} shows performances without the cycle loss, without Chamfer loss, and without any specific triplet sampling strategy during training, simply selecting random shapes.

Table~\ref{tab:ablation} shows that the cycle consistency loss is critical to the success of our method (relative drop of $23\%$ in IoU). Training without Chamfer distance as a reconstruction loss performs slightly better than the identity baseline and $3\%$ below our approach. This highlight the fact that the cycle consistency loss also acts as a reconstruction loss. Finally, our triplet sampling strategy during training provides a small boost.

\begin{table}[!h]
\vskip 0.1in
\begin{center}
\resizebox{\columnwidth}{!}{
\begin{small}
\begin{tabular}{lcccr}
\toprule
 Car/100 shots & Nearest Neighbor & Oracle \\
\midrule
(a) Identity  & 67.60 & 73.59  \\
\midrule
(b) Ours  & \textbf{68.19} & \textbf{75.87}  \\
(c) Ours w/o cycle loss & 52.78 & 59.63 \\
(d) Ours w/o chamfer & 66.21 & 74.31 \\
(e) Ours w/o knn restriction & 67.70 & 75.23 \\
\midrule
\bottomrule
\end{tabular}
\end{small}
}
\end{center}
\caption{\textbf{Ablation Study:}. Given a target shape $T$ and 100 segmented train samples, we select $T$'s nearest neighbors $S$ (1st column), and the oracle source shape which maximizes performances for our approach .  (2nd column). We compare  \textbf{(a)} the identity baseline, with \textbf{(b)} our approach, trained without label supervision,  and \textbf{(c, d, e)} its ablations. The mean IoU is reported. Results are computed on the Car category.}
\label{tab:ablation}
\end{table}

\subsection{Hyperparameter Study}

\begin{figure}[h!]
\centering
\includegraphics[width=9cm, height=3cm]{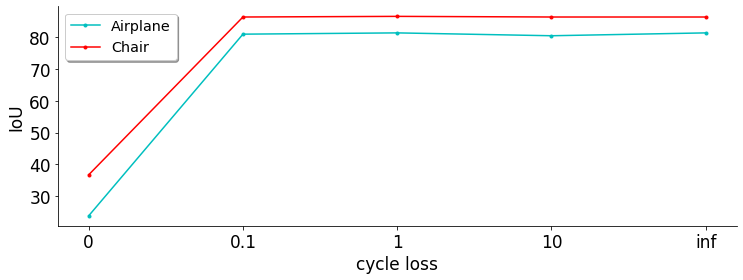} \\
\caption{\final{\textbf{Hyperparameter study. } Study of the influence of the cycle consistency loss from not having it (absciss point "0") to having only the cycle loss (absciss point "inf"). For each target shape, we use the Nearest Neighbors (see ~\ref{sec:segmentation}) criterion to select sources from the full training set. A voting strategy is used on the top 10 source shapes. The mean IoU is reported}}
 \label{fig:hyper_paramater}
\end{figure}

Figure~\ref{fig:hyper_paramater} demonstrates once more that the cycle-consistency loss is the pivotal insight of our method. It also outlines the stability of the results for different weightings of our losses. Note how performances are maintained even in the extreme case with only the cycle-consistency loss. Indeed, the identity function is not a trivial minimum of the cycle consistency loss because of the projection step.

\section{Conclusion}

We have presented a method for learning a parametric transformation between two surfaces that leverages cycle-consistency as a supervisory signal to predict meaningful correspondences. Our method does not require an object template, can operate without any inter-shape correspondences supervision, and does not assume the deformation is nearly isometric. We demonstrate that our method is able to transfer segmentation labels from a very small number of labeled examples significantly better than state-of-the-art methods, and match the segmentation performance when a larger training dataset is provided.

We believe that the large gap between our performance and the ``oracle shape'' which provides maximal accuracy shows that using learned deformations to transfer labels, investigating ways to better understand what source models should be selected and new ways to aggregate information across multiple sources is a very promising research direction. %

{\small
\bibliographystyle{eg-alpha}
\bibliography{new}

\newcommand{\etalchar}[1]{$^{#1}$}
\begin{thebibliography}{\uppercase{vKZHCO11}}

\bibitem[ACBCO17]{Azencot2017}
\textsc{Azencot O., Corman E., Ben-Chen M., Ovsjanikov M.}:
\newblock Consistent functional cross field design for mesh quadrangulation.
\newblock \emph{ACM Trans. Graph. 36}, 4 (July 2017), 92:1--92:13.

\bibitem[BBK06]{Bronstein1168}
\textsc{Bronstein A.~M., Bronstein M.~M., Kimmel R.}:
\newblock Generalized multidimensional scaling: A framework for
  isometry-invariant partial surface matching.
\newblock \emph{Proceedings of the National Academy of Sciences 103}, 5 (2006),
  1168--1172.

\bibitem[BM92]{Besl:1992:MRS:132013.132022}
\textsc{Besl P.~J., McKay N.~D.}:
\newblock A method for registration of 3-d shapes.
\newblock \emph{IEEE Trans. Pattern Anal. Mach. Intell. 14}, 2 (Feb. 1992),
  239--256.

\bibitem[BMRB16]{BoscainiMRB16}
\textsc{Boscaini D., Masci J., Rodol{\`{a}} E., Bronstein M.~M.}:
\newblock Learning shape correspondence with anisotropic convolutional neural
  networks.
\newblock \emph{CoRR abs/1605.06437} (2016).

\bibitem[BR07]{Brown07}
\textsc{Brown B., Rusinkiewicz S.}:
\newblock Global non-rigid alignment of {3-D} scans.
\newblock \emph{ACM Transactions on Graphics (Proc. SIGGRAPH) 26}, 3 (Aug.
  2007).

\bibitem[CFG{\etalchar{*}}15]{ChangFGHHLSSSSX15}
\textsc{Chang A.~X., Funkhouser T.~A., Guibas L.~J., Hanrahan P., Huang Q., Li
  Z., Savarese S., Savva M., Song S., Su H., Xiao J., Yi L., Yu F.}:
\newblock Shapenet: An information-rich 3d model repository.
\newblock \emph{CoRR abs/1512.03012} (2015).

\bibitem[CK15]{Chen15}
\textsc{Chen Q., Koltun V.}:
\newblock Robust nonrigid registration by convex optimization.
\newblock \emph{ICCV} (2015).

\bibitem[EBC17]{Ezuz2017}
\textsc{Ezuz D., Ben-Chen M.}:
\newblock Deblurring and denoising of maps between shapes.
\newblock \emph{Comput. Graph. Forum 36}, 5 (Aug. 2017), 165--174.

\bibitem[GFK{\etalchar{*}}18a]{groueix2018b}
\textsc{Groueix T., Fisher M., Kim V.~G., Russell B., Aubry M.}:
\newblock 3d-coded : 3d correspondences by deep deformation.
\newblock In \emph{ECCV} (2018).

\bibitem[GFK{\etalchar{*}}18b]{groueix2018}
\textsc{Groueix T., Fisher M., Kim V.~G., Russell B., Aubry M.}:
\newblock {AtlasNet: A Papier-M\^ach\'e Approach to Learning 3D Surface
  Generation}.
\newblock In \emph{Proceedings IEEE Conf. on Computer Vision and Pattern
  Recognition (CVPR)} (2018).

\bibitem[HAWG08]{HAWG2008NRUID}
\textsc{Huang Q., Adams B., Wicke M., Guibas L.~J.}:
\newblock Non-rigid registration under isometric deformations.
\newblock In \emph{Computer Graphics Forum} (2008), vol.~27, pp.~1449--1457.

\bibitem[HG13]{Huang2013}
\textsc{Huang Q.-X., Guibas L.}:
\newblock Consistent shape maps via semidefinite programming.
\newblock In \emph{Proceedings of the Eleventh Eurographics/ACMSIGGRAPH
  Symposium on Geometry Processing} (Aire-la-Ville, Switzerland, Switzerland,
  2013), SGP '13, Eurographics Association, pp.~177--186.

\bibitem[HKC{\etalchar{*}}18]{Huang18}
\textsc{Huang H., Kalogerakis E., Chaudhuri S., Ceylan D., Kim V.~G., Yumer
  E.}:
\newblock Learning local shape descriptors from part correspondences with
  multi-view convolutional networks.
\newblock \emph{Transactions on Graphics} (2018).

\bibitem[HZG{\etalchar{*}}12]{Huang2012}
\textsc{Huang Q.-X., Zhang G.-X., Gao L., Hu S.-M., Butscher A., Guibas L.}:
\newblock An optimization approach for extracting and encoding consistent maps
  in a shape collection.
\newblock \emph{ACM Trans. Graph. 31}, 6 (Nov. 2012), 167:1--167:11.

\bibitem[JSZ{\etalchar{*}}15]{jaderberg2015spatial}
\textsc{Jaderberg M., Simonyan K., Zisserman A., et~al.}:
\newblock Spatial transformer networks.
\newblock In \emph{Advances in neural information processing systems} (2015),
  pp.~2017--2025.

\bibitem[KAMC17]{Kalogerakis:2017:ShapePFCN}
\textsc{Kalogerakis E., Averkiou M., Maji S., Chaudhuri S.}:
\newblock 3{D} shape segmentation with projective convolutional networks.
\newblock In \emph{Proc. IEEE Computer Vision and Pattern Recognition (CVPR)}
  (2017).

\bibitem[KB14]{kingma2014adam}
\textsc{Kingma D.~P., Ba J.}:
\newblock Adam: A method for stochastic optimization.
\newblock \emph{arXiv preprint arXiv:1412.6980} (2014).

\bibitem[KHS10]{Kalogerakis:2010:labelMeshes}
\textsc{Kalogerakis E., Hertzmann A., Singh K.}:
\newblock {L}earning {3}{D} {M}esh {S}egmentation and {L}abeling.
\newblock \emph{ACM Transactions on Graphics 29}, 3 (2010).

\bibitem[KLF11]{Kim11}
\textsc{Kim V.~G., Lipman Y., Funkhouser T.}:
\newblock Blended intrinsic maps.
\newblock \emph{Transactions on Graphics (Proc. of SIGGRAPH)}, 4 (2011).

\bibitem[KLM{\etalchar{*}}12]{Kim12}
\textsc{Kim V.~G., Li W., Mitra N.~J., DiVerdi S., Funkhouser T.}:
\newblock {Exploring Collections of 3D Models using Fuzzy Correspondences}.
\newblock \emph{Transactions on Graphics (Proc. of SIGGRAPH)}, 4 (2012).

\bibitem[KTEM18]{cmrKanazawa18}
\textsc{Kanazawa A., Tulsiani S., Efros A.~A., Malik J.}:
\newblock Learning category-specific mesh reconstruction from image
  collections.

\bibitem[LRR{\etalchar{*}}17]{LitanyRRBB17}
\textsc{Litany O., Remez T., Rodol{\`{a}} E., Bronstein A.~M., Bronstein
  M.~M.}:
\newblock Deep functional maps: Structured prediction for dense shape
  correspondence.
\newblock \emph{CoRR abs/1704.08686} (2017).

\bibitem[LSD{\etalchar{*}}18]{Li2018}
\textsc{Li L., Sung M., Dubrovina A., Yi L., Guibas L.~J.}:
\newblock Supervised fitting of geometric primitives to 3d point clouds.
\newblock \emph{CVPR} (2018).

\bibitem[LSP08]{li08global}
\textsc{Li H., Sumner R.~W., Pauly M.}:
\newblock Global correspondence optimization for non-rigid registration of
  depth scans.
\newblock \emph{Computer Graphics Forum (Proc. SGP'08) 27}, 5 (July 2008).

\bibitem[MWZ{\etalchar{*}}14]{Mitra14}
\textsc{Mitra N.~J., Wand M., Zhang H., Cohen-Or D., Kim V.~G., Huang Q.-X.}:
\newblock {Structure-Aware Shape Processing}.
\newblock \emph{SIGGRAPH Course notes} (2014).

\bibitem[MZC{\etalchar{*}}19]{mo2018partnet}
\textsc{Mo K., Zhu S., Chang A., Yi L., Tripathi S., Guibas L., Su H.}:
\newblock {PartNet}: A large-scale benchmark for fine-grained and hierarchical
  part-level {3D} object understanding.

\bibitem[NBCW{\etalchar{*}}11]{nguyen2011optimization}
\textsc{Nguyen A., Ben-Chen M., Welnicka K., Ye Y., Guibas L.}:
\newblock An optimization approach to improving collections of shape maps.
\newblock In \emph{Computer Graphics Forum} (2011), vol.~30, Wiley Online
  Library, pp.~1481--1491.

\bibitem[OBCS{\etalchar{*}}12]{Ovsjanikov2012}
\textsc{Ovsjanikov M., Ben-Chen M., Solomon J., Butscher A., Guibas L.}:
\newblock Functional maps: A flexible representation of maps between shapes.
\newblock \emph{ACM Trans. Graph. 31}, 4 (July 2012), 30:1--30:11.

\bibitem[OMMG10]{OvsjanikovMMG10}
\textsc{Ovsjanikov M., Mérigot Q., Mémoli F., Guibas L.~J.}:
\newblock One point isometric matching with the heat kernel.
\newblock \emph{Comput. Graph. Forum 29}, 5 (2010), 1555--1564.

\bibitem[QSMG16]{qi2016pointnet}
\textsc{Qi C.~R., Su H., Mo K., Guibas L.~J.}:
\newblock Pointnet: Deep learning on point sets for 3d classification and
  segmentation.
\newblock \emph{arXiv preprint arXiv:1612.00593} (2016).

\bibitem[QYSG17]{qi2017pointnetplusplus}
\textsc{Qi C.~R., Yi L., Su H., Guibas L.~J.}:
\newblock Pointnet++: Deep hierarchical feature learning on point sets in a
  metric space.
\newblock \emph{arXiv preprint arXiv:1706.02413} (2017).

\bibitem[RL01]{Rusinkiewicz01}
\textsc{{Rusinkiewicz} S., {Levoy} M.}:
\newblock Efficient variants of the icp algorithm.
\newblock In \emph{Proceedings Third International Conference on 3-D Digital
  Imaging and Modeling} (2001).

\bibitem[ROA{\etalchar{*}}13]{Rustamov2013}
\textsc{Rustamov R.~M., Ovsjanikov M., Azencot O., Ben-Chen M., Chazal F.,
  Guibas L.}:
\newblock Map-based exploration of intrinsic shape differences and variability.
\newblock \emph{ACM Trans. Graph. 32}, 4 (July 2013), 72:1--72:12.

\bibitem[RPWO18]{Ren2018}
\textsc{Ren J., Poulenard A., Wonka P., Ovsjanikov M.}:
\newblock Continuous and orientation-preserving correspondences via functional
  maps.
\newblock \emph{ACM Trans. Graph. 37}, 6 (Dec. 2018), 248:1--248:16.

\bibitem[vKZHCO11]{vankaick11correspsurvey}
\textsc{van Kaick O., Zhang H., Hamarneh G., Cohen-Or D.}:
\newblock A survey on shape correspondence.
\newblock \emph{Computer Graphics Forum 30}, 6 (2011), 1681--1707.

\bibitem[WHC{\etalchar{*}}16]{wei2016dense}
\textsc{Wei L., Huang Q., Ceylan D., Vouga E., Li H.}:
\newblock Dense human body correspondences using convolutional networks.
\newblock In \emph{Computer Vision and Pattern Recognition (CVPR)} (2016).

\bibitem[WSL{\etalchar{*}}18]{Wang18}
\textsc{Wang Y., Sun Y., Liu Z., Sarma S.~E., Bronstein M.~M., Solomon J.~M.}:
\newblock Dynamic graph {CNN} for learning on point clouds.
\newblock \emph{CoRR abs/1801.07829} (2018).

\bibitem[WZL{\etalchar{*}}18]{wang2018pixel2mesh}
\textsc{Wang N., Zhang Y., Li Z., Fu Y., Liu W., Jiang Y.-G.}:
\newblock Pixel2mesh: Generating 3d mesh models from single rgb images.
\newblock In \emph{ECCV} (2018).

\bibitem[WZS{\etalchar{*}}19]{wang2019shape2motion}
\textsc{Wang X., Zhou B., Shi Y., Chen X., Zhao Q., Xu K.}:
\newblock Shape2motion: Joint analysis of motion parts and attributes from 3d
  shapes.
\newblock In \emph{CVPR} (2019), p.~to appear.

\bibitem[YKC{\etalchar{*}}16]{Yi16}
\textsc{Yi L., Kim V.~G., Ceylan D., Shen I.-C., Yan M., Su H., Lu C., Huang
  Q., Sheffer A., Guibas L.}:
\newblock A scalable active framework for region annotation in 3d shape
  collections.
\newblock \emph{SIGGRAPH Asia} (2016).

\bibitem[YLZ{\etalchar{*}}19]{yu2019partnet}
\textsc{Yu F., Liu K., Zhang Y., Zhu C., Xu K.}:
\newblock Partnet: A recursive part decomposition network for fine-grained and
  hierarchical shape segmentation.
\newblock In \emph{CVPR} (2019), p.~to appear.

\bibitem[Zha94]{Zhang94iterativepoint}
\textsc{Zhang Z.}:
\newblock Iterative point matching for registration of free-form curves and
  surfaces, 1994.

\bibitem[ZKA{\etalchar{*}}16]{zhou2016learning}
\textsc{Zhou T., Kr{\"a}henb{\"u}hl P., Aubry M., Huang Q., Efros A.~A.}:
\newblock Learning dense correspondence via 3d-guided cycle consistency.
\newblock In \emph{Computer Vision and Pattern Recognition (CVPR)} (2016).

\bibitem[ZPIE17]{CycleGAN2017}
\textsc{Zhu J.-Y., Park T., Isola P., Efros A.~A.}:
\newblock Unpaired image-to-image translation using cycle-consistent
  adversarial networks.
\newblock In \emph{Computer Vision (ICCV), 2017 IEEE International Conference
  on} (2017).

\bibitem[ZPK18]{Zhou2018}
\textsc{Zhou Q.-Y., Park J., Koltun V.}:
\newblock {Open3D}: {A} modern library for {3D} data processing.
\newblock \emph{arXiv:1801.09847} (2018).

\bibitem[ZSCO{\etalchar{*}}08]{zhang_sgp08}
\textsc{Zhang H., Sheffer A., Cohen-Or D., Zhou Q., van Kaick O., Tagliasacchi
  A.}:
\newblock Deformation-drive shape correspondence.
\newblock \emph{Computer Graphics Forum (Special Issue of Symposium on Geometry
  Processing) 27}, 5 (2008), 1431--1439.

\end{thebibliography}
}

\end{document}